\documentclass[10pt,journal,twoside]{IEEEtran}
    
\newcommand{\etal}{\textit{et al}.}
\newcommand{\ie}{\textit{i}.\textit{e}.}
\newcommand{\eg}{\textit{e}.\textit{g}.}

\newcommand{\zjh}[1]{{\color{black}#1}}

\newcommand{\revise}[1]{{\color{black}#1}}

\newcommand{\newtxt}[1]{{\color{black}#1}}

\usepackage{amsmath,amsfonts}
\usepackage{algorithmic}
\usepackage{algorithm}
\usepackage{array}
\usepackage{booktabs}
\usepackage{textcomp}
\usepackage{stfloats}
\usepackage{url}
\usepackage{verbatim}
\usepackage{graphicx}
\usepackage{cite}
\usepackage{enumitem}

\usepackage[numbers,sort&compress]{natbib}
\usepackage{bbm}
\usepackage{amsmath}
\usepackage{amsfonts}
\usepackage{algorithmic}
\usepackage{algorithm}
\usepackage{array}
\usepackage{graphicx}

\usepackage{subfigure}
\usepackage{hyperref}
\usepackage{makecell}
\usepackage{balance}
\usepackage{booktabs}
\usepackage{color}
\usepackage{multirow}
\usepackage{bbding}
\usepackage{xcolor}
\usepackage{tabu}
\usepackage{lineno}
\usepackage{bm}
\usepackage{mathtools}

\setcitestyle{numbers,sort&compress}
\begin{document}

\title{S$^{5}$Mars: Semi-Supervised Learning for Mars Semantic Segmentation}

\author{Jiahang~Zhang*,
    Lilang~Lin*,
	Zejia~Fan,
	Wenjing~Wang,
      Jiaying~Liu,~\IEEEmembership{Senior Member,~IEEE}

\thanks{This work was supported in part by the National Key Research and Development Program of China under Grant No. 2018AAA0102702, and in part by the National Natural Science Foundation of China under Grant 62172020. \textit{(Corresponding author: Jiaying Liu.)}

*\,Equal contribution. The authors are with the Wangxuan Institute of Computer Technology, Peking University, Beijing, 100080, China, e-mail: \{zjh2020, linlilang, zejia, daooshee, 
liujiaying\}@pku.edu.cn.}

}

 \markboth{IEEE TRANSACTIONS ON GEOSCIENCE AND REMOTE SENSING}{Zhang \MakeLowercase{\textit{et al.}}: S$^{5}$Mars: Semi-Supervised Learning for Mars Semantic Segmentation}

\maketitle
\begin{abstract}
\newtxt{
Deep learning has become a powerful tool for Mars exploration. Mars terrain semantic segmentation is an important Martian vision task, which is the base of rover autonomous planning and safe driving. However, there is a lack of sufficient detailed and high-confidence data annotations, which are exactly required by most deep learning methods to obtain a good model. 
To address this problem, we propose our solution from the perspective of joint data and method design.
We first present a new dataset S$^{5}$Mars for \textit{S}emi-\textit{S}upervi\textit{S}ed learning on Mars \textit{S}emantic \textit{S}egmentation, which contains 6K high-resolution images and is sparsely annotated based on confidence, ensuring the high quality of labels.
Then to learn from this sparse data, we propose a semi-supervised learning (SSL) framework for Mars image semantic segmentation, to learn representations from limited labeled data. Different from the existing SSL methods which are mostly targeted at the Earth image data, our method takes into account Mars data characteristics.
Specifically, we first investigate the impact of current widely used natural image augmentations on Mars images. Based on the analysis, we then proposed two novel and effective augmentations for SSL of Mars segmentation, \textit{AugIN} and \textit{SAM-Mix}, which serve as strong augmentations to boost the model performance. Meanwhile, to fully leverage the unlabeled data, we introduce a soft-to-hard consistency learning strategy, learning from different targets based on prediction confidence. Experimental results show that our method can outperform state-of-the-art SSL approaches remarkably. Our proposed dataset is available at \url{https://jhang2020.github.io/S5Mars.github.io/}.
}
\end{abstract}

\begin{IEEEkeywords}
Mars vision tasks, terrain segmentation, image semantic segmentation, semi-supervised learning.
\end{IEEEkeywords}

\section{Introduction}
\label{sec:intro}
\IEEEPARstart{H}{umans} have shown great enthusiasm for Mars. The history of human research on Mars can date back to the 1960s. So far, more than 30 rovers have been dispatched to the red planet, and the increasing amount of available data promotes the application and development of deep learning algorithms. Deep-learning-based methods have already assisted in prioritizing data selection~\cite{qiu2020scoti}, collecting data, and analyzing data~\cite{goesmann2017mars,dapoian2021science,priyadarshini2021mars}. This paper explores the task of Mars terrain semantic segmentation, which aims to identify the drivable areas and the specific terrains from images. It is of great significance to obstacle avoidance, traversability estimation, data collection, and path planning~\cite{gonzalez2018deepterramechanics,dimastrogiovanni2020terrain}, ensuring the safety and productivity of ongoing and future missions to Mars. 

\zjh{Mars semantic segmentation faces problems from both data and method design. First, the lack of satisfactory and available data hinders the development of deep learning methods to some extent. 
On the one hand, because of the high cost of Mars rovers, limited bandwidth, and data transmission loss from Mars to Earth, collecting Martian data is very expensive. On the other hand, due to the complexity and similarity of the terrain, delicate and dense pixel-level labeling is highly specialized and time-consuming. 
Accordingly, previous datasets~\cite{schwenzer2019labelmars,AI4Mars} are not satisfactory because of the low-quality annotations or the roughly defined categories. AI4Mars~\cite{AI4Mars}, a newly published Mars terrain segmentation dataset, only defines four simple categories which are difficult to meet the actual requirements of complex terrain identification. Besides, some datasets~\cite{schwenzer2019labelmars,AI4Mars} collected through crowdsourcing often do not have satisfactory annotation quality due to inconsistent standards. 
}

\newtxt{From a methodological point of view, the existing methods heavily rely on large amounts of training data and lack targeted and effective design. Early works directly applied a certain machine learning algorithm such as Support Vector Machines (SVM)~\cite{dimastrogiovanni2020terrain}. With the rapid development of deep learning, the terrain segmentation performance is greatly improved by methods based on deep neural networks~\cite{gonzalez2018deepterramechanics,AI4Mars,rothrock2016spoc}. 
However, they still rely on fully supervised learning pipelines that require a lot of high-quality labeled data, which is often difficult to achieve. To this end, semi-supervised learning (SSL) has attracted lots of attention, which learns representations from limited labeled data as well as the amounts of unlabeled data. \revise{However, most existing SSL methods are designed for Earth image data and cannot be directly transferred to Mars image segmentation tasks, due to the properties of Mars images. First, the color of Mars images is less diverse. Traditional color augmentations, which are crucial and widely used in SSL works~\cite{sohn2020fixmatch,zhao2023augmentation,yang2023revisiting}, can cause over-distortion problem~\cite{yuan2021simple} in the form of color distribution shift for the Mars images, and fail to improve the performance as shown in Fig.~\ref{fig:aug_comp}. Note that color distribution shift can arise in different data domains with some similar properties as the Mars images, \eg, less diverse color distributions, when applying the traditional color augmentations. Nevertheless, it is still less explored in previous SSL works, especially from the perspective of data augmentations. Besides, the objects in Mars images are often with irregular contours and obvious occlusions, \eg, between rocks and soil/sand. As a result, the high background complexity makes the model suffer from greater uncertainty in the consistency learning of unlabeled data, leading to the sub-optimal representations. Moreover, some categories are more confusing between each other, \eg, rocks and bedrocks, soil and sand, which require more fine-grained representations to distinguish.}
}

In summary, there are two main challenges in the Mars terrain segmentation task: 1) the lack of data with adequate detailed and high-confidence annotations, 2) insufficient studies targeted at SSL on Mars image data.
\zjh{We solve the above problems from the perspective of \textit{both data and method design}, which are named \textbf{S}emi-\textbf{S}upervi\textbf{S}ed \textbf{S}emantic \textbf{S}egmentation for \textbf{Mars} (\textbf{S$^{5}$Mars}). We first create a new dataset to provide a high-quality and fine-grained labeled data for Mars terrain segmentation. Our dataset contains 6K high-resolution images captured on the surface of Mars, each of which is annotated by a professional team. There are 9 categories defined in our dataset, including sky, ridge, soil, sand, bedrock, rock, rover, trace, and hole, respectively. To improve the quality of labels, the annotation of the dataset adopts a sparse labeling style, \ie, only the area with high human confidence is annotated. 
}

\newtxt{To learn from this sparse data, we propose a new semi-supervised framework for Mars image terrain segmentation. Our method is based on the recently popular consistency regularization-based methods, which utilize weak-to-strong augmentations to generate the perturbation while pursuing the perturbation consistency. Specifically, we first investigate the impact of widely used Earth image augmentations on Mars data and are surprised to find their adverse effects on the SSL of Mars segmentation. Based on this analysis, we further propose two novel 
and effective augmentations, \textit{AugIN} and \textit{SAM-Mix}. \textit{AugIN} exchange statistics between images to generate new data views while avoiding drastic color distribution shift. \textit{SAM-Mix} utilizes the pretrained Segment-Anything Model (SAM)~\cite{kirillov2023segment} to generate high-quality object masks, reducing the uncertainty of the mixed images. These two data augmentations lead to better consistency learning and improve the performance remarkably.
Finally, we introduce the soft-to-hard consistency learning strategy, which utilizes the soft pseudo-labels in low-confidence regions, while using the hard pseudo-labels in high-confidence regions, fully taking advantage of the unlabeled data.
Extensive experiments and ablation studies verify the effectiveness of the proposed method.}

\zjh{Our contributions can be summed up as follows:
\begin{itemize}
    \item We collect a new fine-grained labeled Mars dataset for terrain semantic segmentation, which contains a large amount of Martian geomorphological data. Our dataset is sparsely annotated by a professional team under multiple rounds of inspection rework. The high-quality dataset can provide accurate and rich segmentation guidance.
    
    \item \newtxt{We systematically study the data augmentations used in current mainstream SSL methods and find their detrimental impact on Mars image segmentation, especially the traditional color augmentations. We analyze this problem and further propose two new and effective augmentations, \textit{SAM-Mix} and \textit{AugIN}, boosting the performance of SSL methods for Mars image segmentation.  }

    \item \newtxt{To fully take advantage of the unlabeled data, a soft-to-hard consistency learning strategy is introduced. The model is constrained to learn consistency by the hard pseudo-labels in high-confidence regions as well as the soft pseudo-labels in low-confidence regions, further improving the consistency.}
\end{itemize}
}

{The rest of this article is organized as follows. In Section~\ref{sec:related}, we provide a detailed survey on Martian datasets and a brief review of deep learning for Mars. Section~\ref{sec:dataset} introduces our proposed Mars segmentation dataset. Then we present our framework for Mars semantic segmentation in Section~\ref{sec:semi}. Experimental results and analysis are shown in Section~\ref{sec:exp}.
The conclusion is finally given in Section~\ref{sec:conclusion}.}

\section{Related Works}
\label{sec:related}

\begin{table*}[]
\centering
\caption{Summary of Mars terrain-aware datasets.}
\label{tab:mars_summary}
\renewcommand\arraystretch{1.2}
{
\setlength{\tabcolsep}{4mm}{
\begin{tabular}{c|c|lllll}
\toprule
\multicolumn{1}{c|}{Type}                                                          & Source                                                                                           & Dataset                                        & Scale               & Classes            & \multicolumn{2}{l}{Description}                                                                                           \\ \hline
\multirow{12}{*}{Real}                                                             & \multirow{10}{*}{Curiosity rover}                                                                 & \multicolumn{1}{l}{\multirow{2}{*}{\cite{rothrock2016spoc}}} & 5k                & -                  & \multicolumn{2}{l}{Wheel slip and slope angles prediction}                                                                \\ \cline{4-7} 
                                                                                   &                                                                                                  & \multicolumn{1}{l}{}                           & 700           & 6                  & \multicolumn{2}{l}{Terrain segmentation}                                                                                  \\ \cline{3-7} 
                                                                                   &                                                                                                  & \cite{gonzalez2018deepterramechanics}                                     & 300                 & 3                  & \multicolumn{2}{l}{Terrain classification}                                                                                \\ \cline{3-7} 
                                                                                   &                                                                                                  & \cite{li2020autonomous}                                      & 620                 & 4                  & \multicolumn{2}{l}{Terrain classification}                                                                                \\ \cline{3-7} 
                                                                                   &                                                                                                  & \cite{WagstaffLSGGP18}                                        & 6k                & 24                 & \multicolumn{2}{l}{Terrain classification}                                                                                \\ \cline{3-7} 
                                                                                   &                                                                                                  & \cite{xiao2021kernel}                                       & 405                 & -                  & \multicolumn{2}{l}{Rock detection}     \\ \cline{3-7} 
                                                                                   &                                                                                                  & \cite{liu2023rockformer}                                       & 8k                & -                & \multicolumn{2}{l}{Rock detection}                                                                                  \\ \cline{3-7} 
                                                                                   &                                                                                                  & \cite{qiu2020scoti}                                       & 1k                & -                  & \multicolumn{2}{l}{Image description}                                                                                     \\ \cline{3-7} 
                                                                                   &                                                                                                  & \cite{kerner2018context}                                       & 310k                & -                  & \multicolumn{2}{l}{\begin{tabular}[c]{@{}l@{}}Compressed image quality evaluation\\ with automatic labeling\end{tabular}} \\ \cline{2-7} 
                                                                                   & Opportunity, Spirit rovers                                                                        & \cite{thompson2007performance}                                       & 117                 & -                  & \multicolumn{2}{l}{Rock detection}                                                                                        \\ \cline{2-7} 
                                                                                   & \multirow{4}{*}{\begin{tabular}[c]{@{}c@{}}Curiosity, Opportunity, \\ Spirit rovers\end{tabular}} & \cite{thompson2012smart}                                       & 46                  & 2                  & \multicolumn{2}{l}{Terrain segmentation}                                                                                  \\ \cline{3-7} 
                                                                                   &                                                                                                  & \cite{AI4Mars}                                       & 35k           & 4                  & \multicolumn{2}{l}{Terrain segmentation} 
                                                                                   
                                                                                 \\ \cline{3-7} 
                                                                                   &                                                                                                  & \cite{li2022stepwise}                                       & 5k           & 9                  & \multicolumn{2}{l}{Terrain segmentation} 
                                                                                   
                                                                                   \\ \cline{3-7} 
                                                                                   &                                                                                                  & \cite{schwenzer2019labelmars}                                       & 5k                & 6 (17 sub)                 & Terrain segmentation  
                                                                                   
                                                                                   \\ \hline
\begin{tabular}[c]{@{}c@{}}Real + Synthetic\end{tabular}                         & Curiosity rover                                                                                  & \cite{wilhelm2020domars16k}                                       & 30k                 & 5                  & \multicolumn{2}{l}{Terrain classification}                                                                                \\ \hline
Synthetic                                                                          & ROAMS rover simulator                                                                            & \cite{thompson2007performance}                                       & 55                  & -                  & \multicolumn{2}{l}{Rock detection}                                                                                        \\ \hline
\multirow{6}{*}{Simulation field}                                                  & \begin{tabular}[c]{@{}c@{}}Atacama Desert\\ Zoë rover prototype\end{tabular}                     & \cite{niekum2005reliable}                                       & 30                  & -                  & \multicolumn{2}{l}{Rock detection}                                                                                        
\\ \cline{2-7} 
                                                                                   & {\begin{tabular}[c]{@{}c@{}}JPL Mars Yard \\FIDO rover Platform \end{tabular}}                & \multirow{1}{*}{\cite{thompson2007performance}}                      & \multirow{1}{*}{35} & \multirow{1}{*}{-} & \multicolumn{2}{l}{\multirow{1}{*}{Rock detection}}                                                                  
                                                                             \\ \cline{2-7} 
                                                                                   &          {\begin{tabular}[c]{@{}c@{}}JPL Mars Yard \\Athena rover Platform\end{tabular}}                                & \cite{higa2019vision}                      & 91k & - & \multicolumn{2}{l}{Rover energy consumption}                                                                
                                                                                   
                                                                                   \\ \cline{2-7} 
                                                                                   & Devon Island                                                                                     & \cite{furlan2019rock}                                       & 400                 & -                  & \multicolumn{2}{l}{Rock detection}        
                                                                                       \\ \hline
\multirow{2}{*}{\begin{tabular}[c]{@{}c@{}}Real +\\ Simulation field\end{tabular}} & \multirow{2}{*}{Opportunity, Spirit rovers}                                                       & \multirow{2}{*}{\cite{xiao2017autonomous}}                      & \multirow{2}{*}{36} & \multirow{2}{*}{2} & \multicolumn{2}{l}{\multirow{2}{*}{Terrain segmentation}}                                                                 \\
                                                                                   &                                                                                                  &                                                 &                     &                    & \multicolumn{2}{l}{}                                                                                                      \\ \bottomrule
\end{tabular}
}}
\end{table*}

\subsection{Deep Learning for Mars}
\zjh{With the increasing amount of available data and the rapid development of computing power, deep learning is playing an increasingly important role in Mars exploration.

For many reasons such as limited computing resources, existing deep learning methods are usually ex-situ (Earth edge). For terrain identification, Deep Mars~\cite{WagstaffLSGGP18} trains an AlexNet to classify engineering-focused rover images (\eg, those of rover wheels and drill holes) and orbital images. However, it can only recognize one object in a single image. The Soil Property and Object Classification (SPOC)~\cite{rothrock2016spoc} proposes to segment the Mars terrains in an image by using a fully convolutional neural network. Swan \etal~\cite{AI4Mars} collect a terrain segmentation dataset and evaluate the performances using DeepLabv3+~\cite{DeeplabV3_plus}. Considering the dependence of existing methods on large amounts of data, \cite{less} utilizes a self-supervised method and trains the model on less labeled images. Recently, Transformer-based network is studied~\cite{liu2023rockformer,xiong2023marsformer} for Martian rock segmentation task. For other tasks, Zhang \etal~\cite{zhang2018novel} deal with Mars visual navigation problem by utilizing a deep neural network, which can find the optimal path to the target point directly from the global Martian environment.

Meanwhile, intrigued by the vision of autonomous probes that rely on deep learning even without human-in-the-loop requirements, scientists are studying the potential of implementing in-situ (Mars edge) deep learning algorithms using high-performance chips~\cite{ono2020maars}. For example, the Scientific Captioning of Terrain Images (SCOTI)~\cite{qiu2020scoti} model automatically creates captions for pictures of the Martian surface based on LSTM, which helps selectively transfer more valuable data within downlink bandwidth limitations. For energy-optimal driving, Higa \etal~\cite{higa2019vision} propose to predict energy consumption from images based on a PNASNet-5~\cite{liu2018progressive}.

\revise{However, many existing works still directly transfer the technology designed for the Earth scene to the Mars task, which can be sub-optimal due to the properties of Mars data. Meanwhile, due to the significant bandwidth and computational resource limitations, the model is expected to be lightweight and efficient, and hence the large models are unsuitable to employ. Most importantly, most of these methods require a lot of annotated training data, which is expensive and hard to obtain. Although some domain adaptation methods~\cite{li2022stepwise, sun2016deep} also can learn the target domain knowledge without many labels, they still suffer from taxonomy
inconsistencies in segmentation detail as discussed in~\cite{less,lambert2020mseg}.} To this end, in this paper we present a powerful semi-supervised learning framework designed for the Mars images.}

\subsection{Datasets for Mars Vision}
Datasets are the basis for intelligent algorithms development. At present, there are various datasets of planetary surfaces, such as digital simulation Lunar landscape segmentation dataset ALLD.
As for Mars, the commonly used terrain-aware datasets can be divided into three categories: rover shooting real data, artificial synthetic data, and earth simulation field shooting data. The rover shooting data are captured by devices of rovers that land on Mars. The number of rovers sent to Mars will gradually increase along with the progress of space research. However, the amount of data available now is still relatively limited. Synthesizing Mars datasets by means of digital modeling simulation or adversarial learning is an important data supplement, but can differ greatly from the real Mars data. Earth simulation field shooting way requires building a simulation platform or finding a similar landscape on Earth to Mars, which is difficult to implement. 
The current Mars terrain-aware datasets are shown in Table~\ref{tab:mars_summary}, which are shot by the Mars rovers. A large proportion of them have an image quantity of less than 1000, which can not meet the training needs of the machine learning models. The richness of Mars terrain-aware datasets still needs to be strengthened.

\begin{figure*}
\centering
	\subfigure[Sky]{
		\begin{minipage}[b]{0.26\textwidth}
		\centering
		\includegraphics[width=1\textwidth]{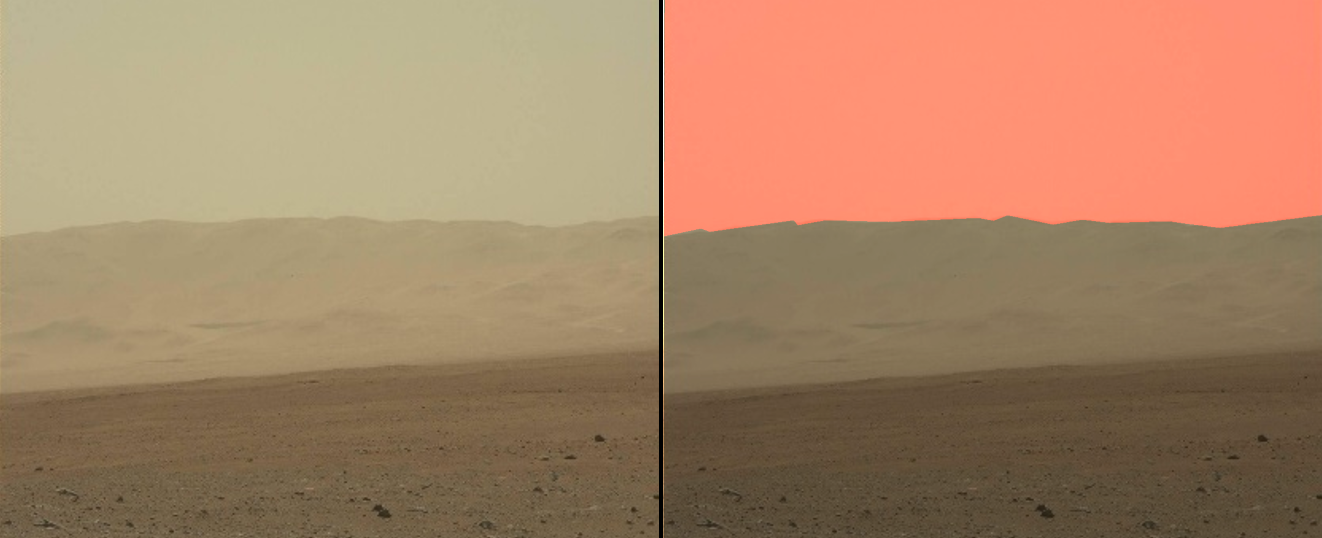} 
		\end{minipage}}
		\hspace{0.4pt}
	\subfigure[Ridge]{
		\begin{minipage}[b]{0.26\textwidth}
  	 	\centering
  	 	\includegraphics[width=1\textwidth]{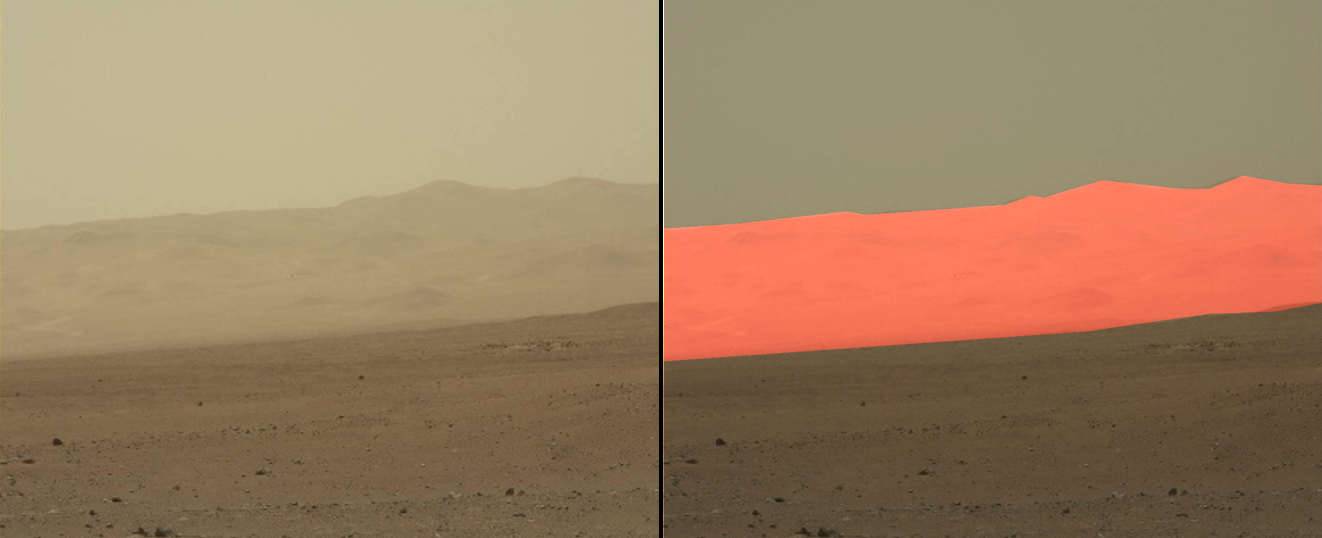}
		\end{minipage}}
		\hspace{0.4pt}
	\subfigure[Soil]{
		\begin{minipage}[b]{0.26\textwidth}
  	 	\centering
			\includegraphics[width=1\textwidth]{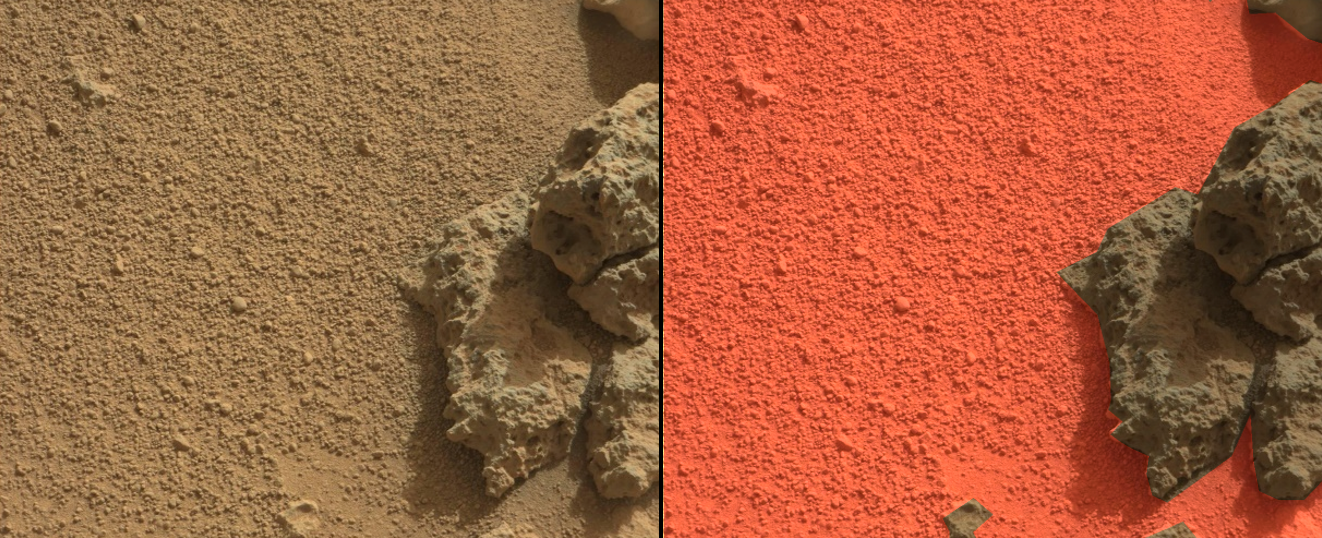} 
		\end{minipage}
	}
	
	\vspace{-5pt}
	\subfigure[Sand]{
		\begin{minipage}[b]{0.26\textwidth}
  	 	\centering
	 	\includegraphics[width=1\textwidth]{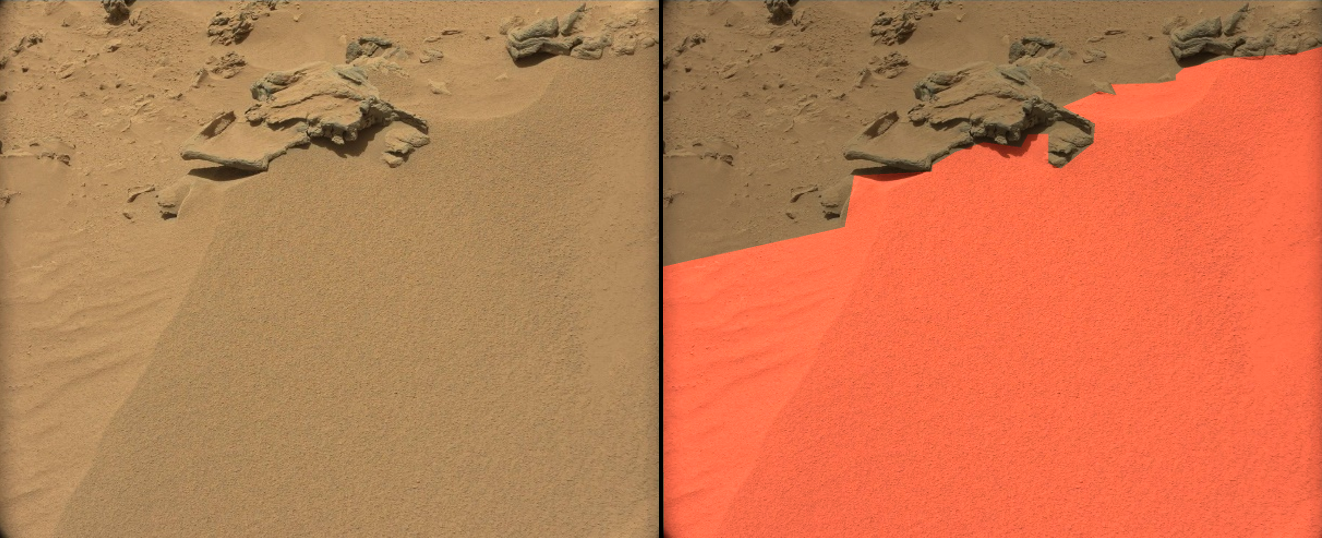}
		\end{minipage}
	}
	\subfigure[Bedrock]{
		\begin{minipage}[b]{0.26\textwidth}
  	 	\centering
	 	\includegraphics[width=1\textwidth]{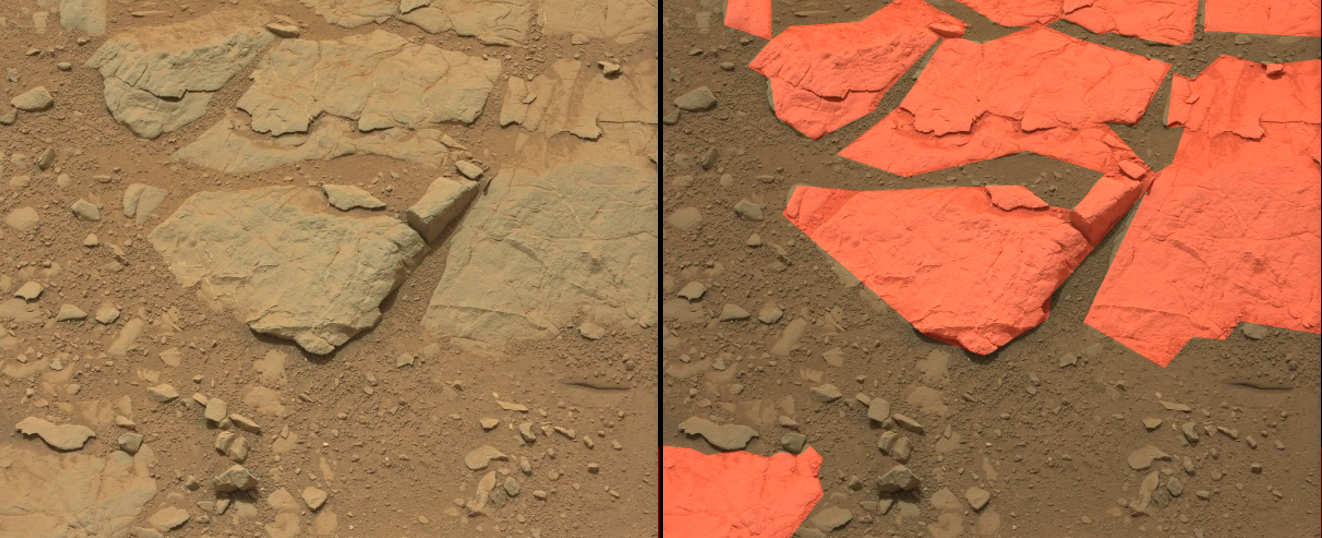}
		\end{minipage}
	}
	\subfigure[Rock]{
		\begin{minipage}[b]{0.26\textwidth}
  	 	\centering
	 	\includegraphics[width=1\textwidth]{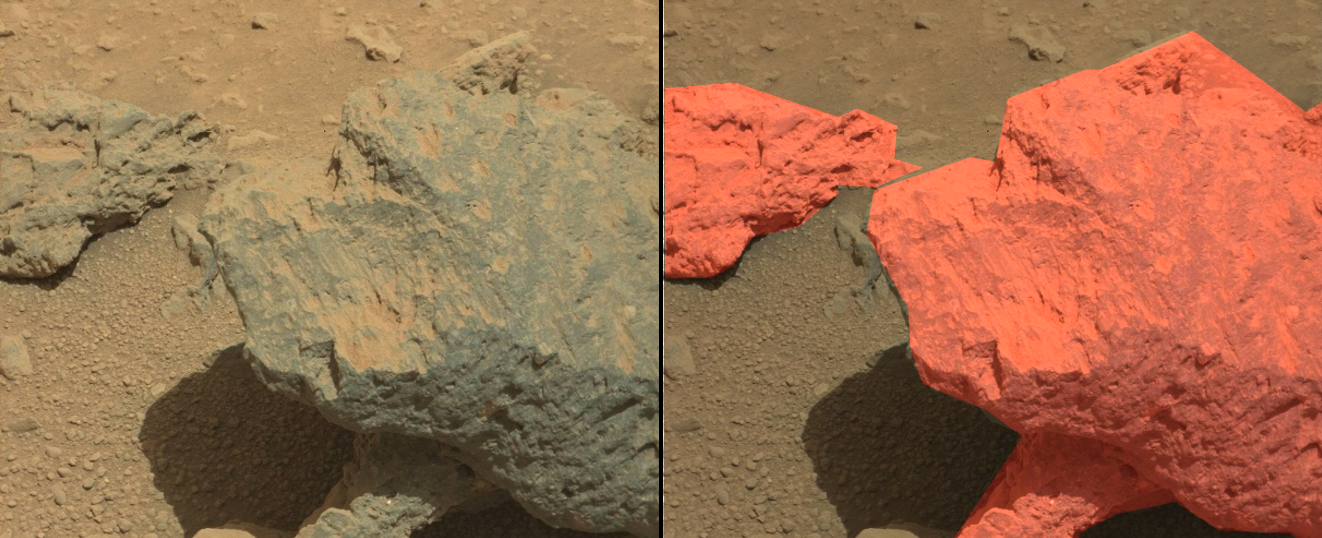}
		\end{minipage}
	}
	
	\vspace{-5pt}
	\subfigure[Rover]{
		\begin{minipage}[b]{0.26\textwidth}
  	 	\centering
	 	\includegraphics[width=1\textwidth]{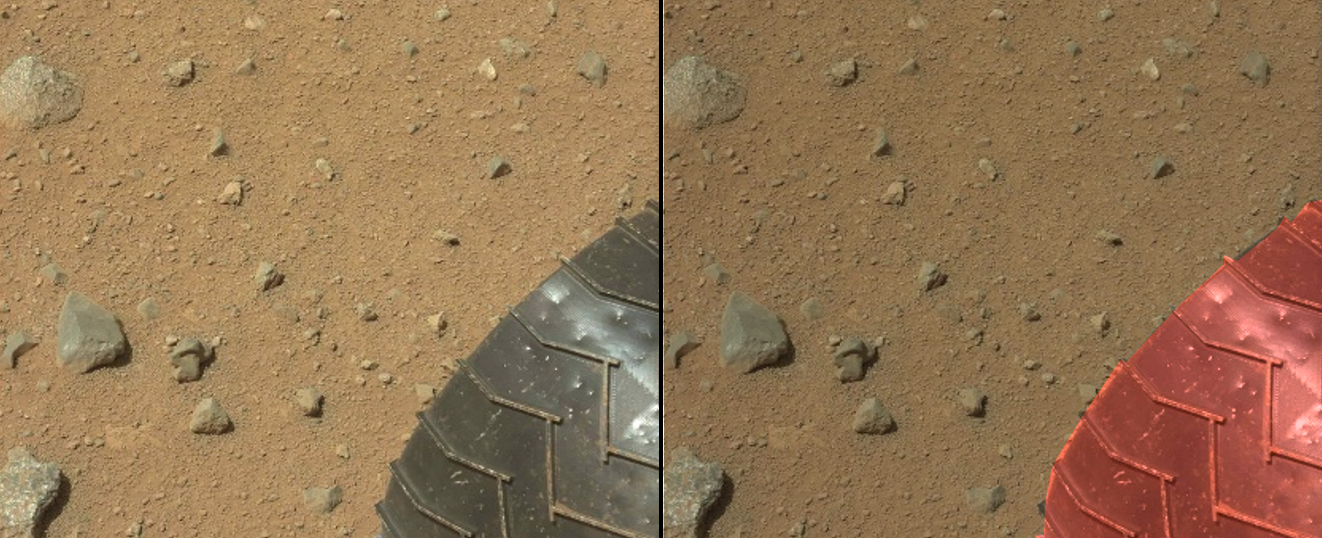}
		\end{minipage}
	}
	\subfigure[Trace]{
		\begin{minipage}[b]{0.26\textwidth}
  	 	\centering
	 	\includegraphics[width=1\textwidth]{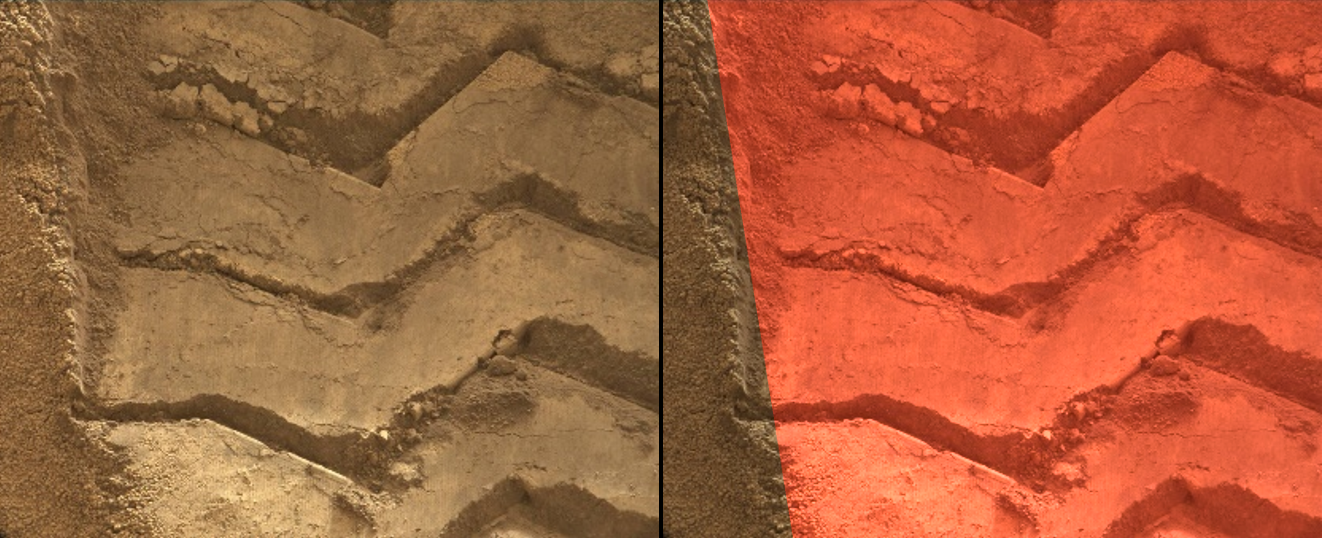}
		\end{minipage}
	}
	\subfigure[Hole]{
		\begin{minipage}[b]{0.26\textwidth}
  	 	\centering
	 	\includegraphics[width=1\textwidth]{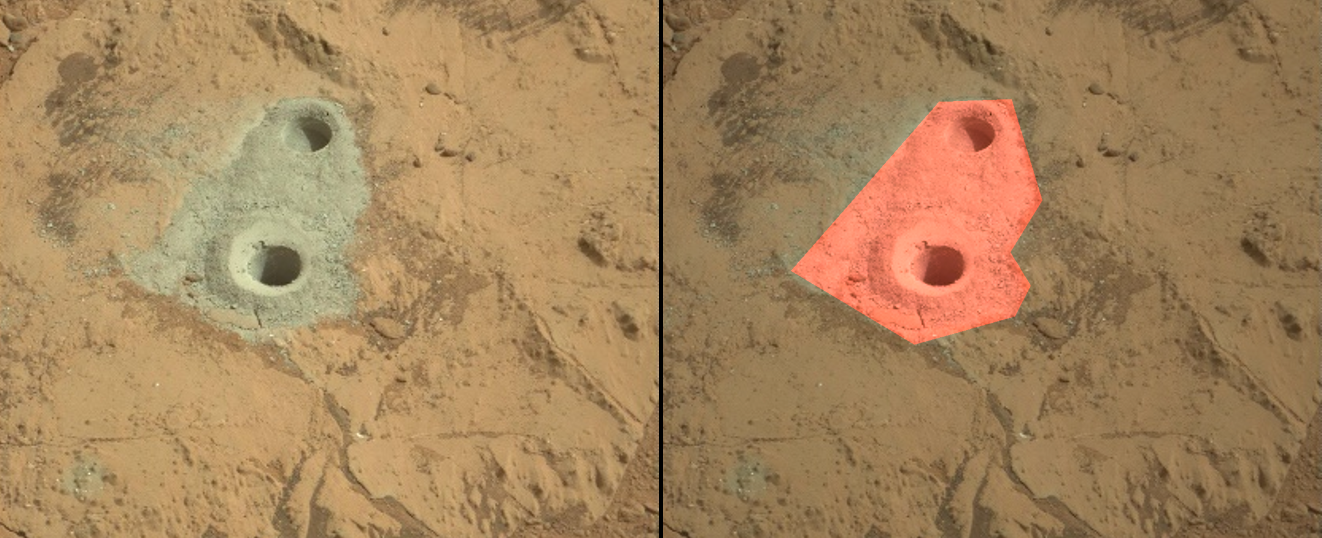}
		\end{minipage}
	}
	\caption{Examples for each label category (highlighted in red).}
	\label{fig:example}
\end{figure*}

\begin{figure}
\color{black}
	\subfigure[\revise{The number of images with $n$ categories appearing simultaneously.}]{
		\begin{minipage}[b]{0.305\textwidth}
			\includegraphics[width=0.9\textwidth]{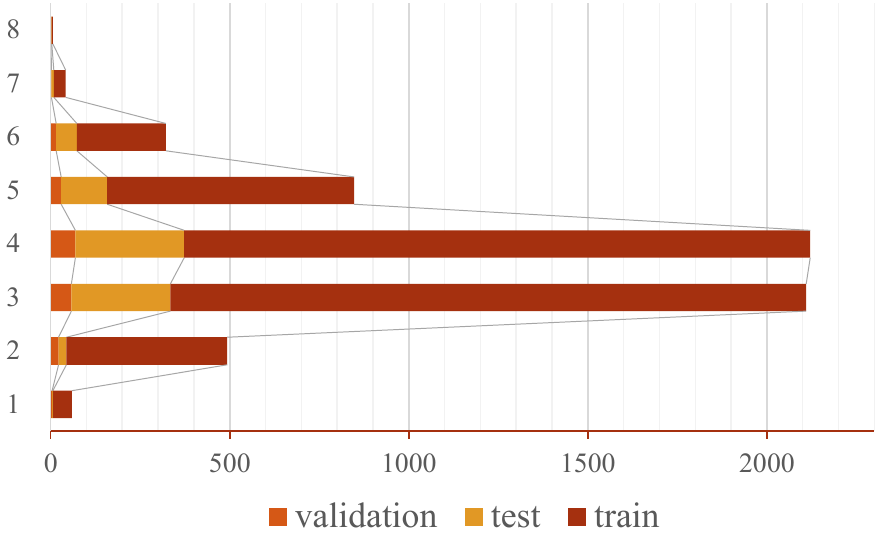}
			\label{fig:number_dis}
		\end{minipage}}
	\subfigure[The distribution of different label area.]{
		\begin{minipage}[b]{0.16\textwidth}
  	 	\includegraphics[width=1\textwidth]{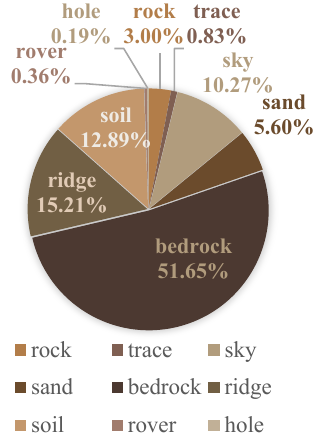}
  	 	\vspace{-3pt}
  	 	\label{fig:area_dis}
		\end{minipage}}
		\caption{Numerical statistics on our S$^{5}$Mars dataset. The figures show the richness of the categories from two aspects: distribution of the number of different labels in each image and distribution of label area. \revise{Note no image contains 9 labels simultaneously in its annotation, so it is omitted in (a).}
 }
	\label{fig:statistic}	
\end{figure}

\subsection{Semi-Supervised Learning}
\newtxt{Semi-supervised learning~\cite{ouali2020overview} utilizes the manifold structure of unlabeled data to assist learning with labeled data. The key issue is how to exploit the information of unlabeled data. Generally, the cross-entropy loss is optimized by the ground-truth label on the labeled data, while a regularization term is applied to the model \textit{w.r.t} the unlabeled data. For example, the pseudo-label method~\cite{xie2020self} assigns pseudo-labels to unlabeled data through a classifier trained on supervised data, which is typical in entropy minimization methods. 

Regarding the utilization of unlabeled data, many researchers have conducted extensive studies, covering unsupervised contrastive learning~\cite{wang2022semi,zhong2021pixel}, uncertainty attention mechanism~\cite{zhong2021pixel,hu2022semi}, and extra correcting networks~\cite{mendel2020semi}. However, these methods improve the performance at the cost of increasing space and computational complexity.
Recently, consistency regularization-based methods have attracted lots of attention, due to their simplicity and effectiveness. They rely on various perturbation techniques (augmentations) to generate different data patterns, which maintain similar semantic information as the original data. Then, the consistency regularization objective is performed to guide the learning of the unlabeled data. MixMatch~\cite{berthelot2019mixmatch} mixes labeled and unlabeled data using MixUp~\cite{MixUp} and performs consistency regularization utilizing low-entropy labels. 
FixMatch~\cite{sohn2020fixmatch} further assigns pseudo-labels, which are the predictions by teacher model on weakly augmented data, to the corresponding strongly augmented data. Inheriting from FixMatch, FlexMatch~\cite{zhang2021flexmatch}
and FreeMatch~\cite{wang2022freematch} propose to learn the threshold for different classes adaptively to filter the low-confidence pseudo-labels. Zhao \etal~\cite{zhao2023augmentation} proposed a series of strong data augmentations to enhance the augmented space. UniMatch~\cite{yang2023revisiting} utilizes both the data-level and feature-level augmentations to constrain the consistency learning.

In these consistency regularization methods, the augmentations, \ie, the perturbation techniques, are crucial for the semantic segmentation. Many techniques, \eg, geometric-based, color-based, mixing-based, and feature perturbation-based methods have been studied. Furthermore, some random auto-augment modules are developed~\cite{cubuk2020randaugment} to further expose data patterns. \revise{However, these methods are less suitable and effective for the Mars semantic segmentation, due to the special properties of Mars images as we discussed in Section~\ref{sec:intro}.} Therefore, it is significant and critical for the study of augmentations for Mars image data.
In this paper, we analyze the characteristics of Mars images, and study the performance of existing common augmentations on Mars data. Meanwhile, we propose two effective new augmentations to boost SSL for Mars image segmentation.

}

\section{Proposed Mars Imagery Segmentation Dataset}
\label{sec:dataset}

\zjh{To solve the problem of scarce available training data for deep learning, }
we create a fine-grained labeled Mars dataset for the exploration on Mars surface, namely, S$^{5}$Mars. Our dataset includes 6,000 high-resolution images taken on the surface of Mars, by color mast camera (Mastcam) from Curiosity (MSL), with the spatial resolution of 1200 $\times$ 1200. 
The dataset is divided in a roughly stratified sampling manner to make the label distribution similar among different splits, yielding a training set of 5000 images, a validation set of 200 images and a test set of 800 images.

\subsection{Labeling Process}

There are 9 label categories, sky, ridge, soil, sand, bedrock, rock, rover, trace, and hole, respectively. Examples of each category are shown in Fig.~\ref{fig:example}. The labeling criteria are as follows:
\begin{itemize}
    \item \textbf{Sky.} The Martian sky, often at the top of a distant image, bounded by the upper edge of a mountain or horizon.
    \item \textbf{Ridge.} The distant peaks bounded by the sky above and the horizon below.
    \item \textbf{Soil.} Unconsolidated or poorly consolidated weathered material on the surface of Mars, with larger and coarse-grained grains containing small stones.
    \item \textbf{Sand.} Granular material, more fluid, less viscous, some with windward and leeward sides, most of the time with sand ridges.
    \item \textbf{Bedrock.} Partially covered by the soil and buried at varying depths.
    \item \textbf{Rock.} A stone that is completely exposed to the ground and is roughly lumpy or oval in shape, usually with distinct shadows.
    \item \textbf{Rover.} The rover itself.
    \item \textbf{Trace.} The trace left by the rover when it passed over the ground.
    \item \textbf{Hole.} The hole left by the rover during its sampling operation on Mars,  contains the surrounding soil of different colors.
    
\end{itemize}

Martian surface condition is complicated due to the harsh and volatile Martian environment. The terrain types can mix and overlap with each other and it becomes hard for humans to distinguish the correct categories clearly. Considering the situation, we apply \textit{sparse labeling}, \ie, only the pixels with enough human confidence are labeled. The overall annotation priority is in a coarse-to-fine manner, which means we label each image in order of object size, \revise{and the total pixel annotation ratio is 48.9\%}. 
As for the annotating process, the annotation rules are discussed more than ten times to keep consistency and preciseness. Each annotation result passes more than two turns of quality inspections. Annotation work is carried out by a professional team, where 90\% of the annotators have been engaged in such annotation work more than six times. The annotation time of each terrain image is about 30 minutes.

\subsection{Comparison and Analysis}

We make a statistical analysis on the semantic labels in the dataset, as shown in Fig.~\ref{fig:statistic}. We show the distribution of the number of different labeled categories contained in each image in Fig.~\ref{fig:number_dis}. Most images are relatively complex with three or four annotations in one scene. This distribution on training, validation and test sets keeps in good consistency. 

We make statistics of the distribution of label area of each category, as shown in Fig.~\ref{fig:area_dis}. The total pixel-wise label ratio is $49\%$. For the labeled regions, bedrock is the label of the largest annotation area, ridge the second.
Rocks appear in most of the images in the dataset, but the total area is small. 
The artificial impact, \eg, rover, trace, and hole, accounts for few portions of the labeled area, but they have a greater variety of shapes and are crucial to the observation and judgment system for intelligence research on Mars.

\begin{figure*}[th]
	\centering
  	 	\includegraphics[width=1\textwidth]{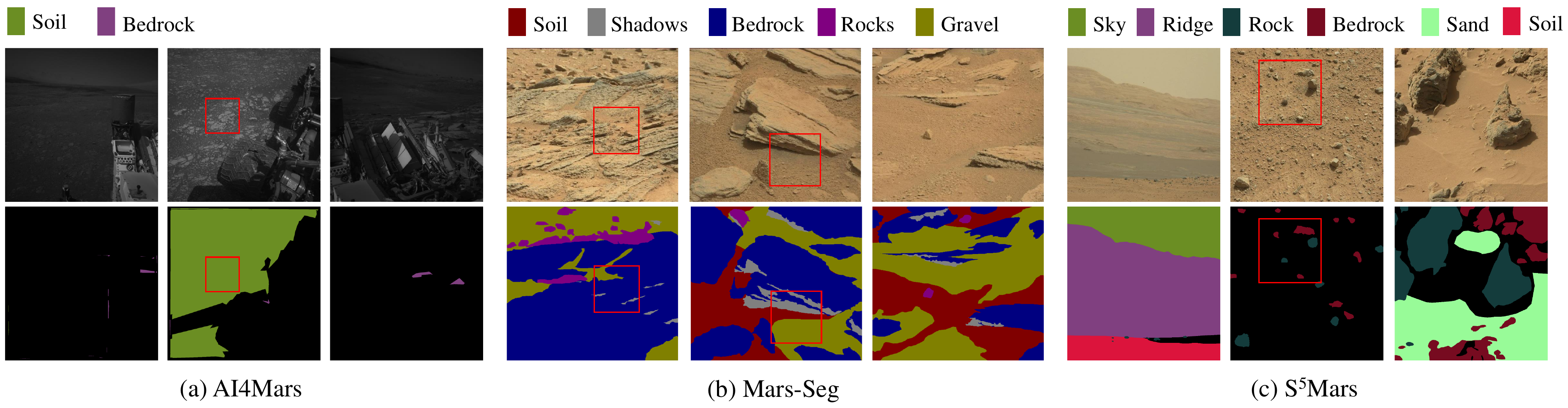}
	\caption{Some image-label examples in different datasets: (a) AI4Mars~\cite{AI4Mars}. Due to the few defined categories, the annotation diversity and adequacy are insufficient. Meanwhile, there are some cases of mislabeling (red box). (b) Mars-Seg~\cite{li2022stepwise}, which gives a complete pixel-level labeling. However, the label can be misleading when different categories mix up with each other (red box). (c) Our dataset S$^{5}$Mars, which provides accurate labeling for regions with high confidence.}
	\label{fig:datsets_example}
\end{figure*}

\newtxt{AI4Mars contains 4 categories with gray-scale images available solely, which can only provide limited task knowledge. Moreover, since AI4Mars is a crowdsourcing project, though the number of submissions is large, the annotators may have inconsistent understandings of labeling standards, which can lead mislabeling in the annotations as shown in Fig.~\ref{fig:datsets_example}(a). In contrast, our dataset is equipped with high-resolution RGB images including 9 semantic categories. Meanwhile, we establish clear labeling criteria and provide professional training to annotators, making the proposed dataset more reliable.}

Mars-Seg~\cite{li2022stepwise} is also a public Mars terrain segmentation dataset. The dataset has 1,064 high-resolution grayscale images and 4,184 RGB images with a spatial resolution of $560 \times 500$, while S$^{5}$Mars is composed of high-resolution RGB images, which offers more accurate and more abundant semantic information for detection and segmentation tasks. Meanwhile, categories in Mars-Seg like gravel, sand, and rocks mix up with each other, making it hard to determine the terrain scene into any one category\newtxt{, as shown in Fig.~\ref{fig:datsets_example}(b)}. Instead, S$^{5}$Mars applies confidence-based sparse-labeled manner. This way we guarantee the labels are strongly representative in each category and reduce the label noise introduced in the labeling work{, as shown in Fig.~\ref{fig:datsets_example}(c)}.

\section{\newtxt{The Proposed Method}}
\label{sec:semi}
In this section, we introduce the proposed method for Mars image semantic segmentation. 
The overview and motivations are first provided in Section~\ref{sec:pre}. Then, we systematically investigate the augmentations for Mars images in Section~\ref{sec:aug}, and propose two effective augmentation techniques based on the analysis. Finally, in Section~\ref{sec:sth}, we introduce the soft-to-hard consistency learning strategy and present the full model.

\subsection{Preliminaries and Motivation}\label{sec:pre}
\textbf{1) Overview:}
As introduced in the previous sections, our proposed dataset is annotated in a sparse style, \ie, some areas of an image are annotated and some are not. For clarity, we no longer distinguish between unlabeled images and unlabeled areas in an image, which can be aligned with a few minor changes. Following the dominant consistency regularization semi-supervised methods~\cite{ouali2020overview,sohn2020fixmatch}, the model is trained on both labeled and unlabeled images simultaneously. Given a batch of labeled images 
$\mathcal{B}_l=\{({\textbf{x}_i}, {\textbf{y}_i})\}_{i=1}^{|\mathcal{B}_l|}$
and a batch of unlabeled images $\mathcal{B}_u=\{(\textbf{u}_i)\}_{i=1}^{|\mathcal{B}_u|}$,
the goal of SSL is to train a model $f(\cdot;\theta)$ with good representations by optimizing the following objective $\mathcal{L}$:
\begin{align}
    \label{equ:loss:total}
    \mathcal{L} = \mathcal{L}_{sup} + \lambda_u \mathcal{L}_{unsup},
\end{align}
where $\mathcal{L}_{sup}$ is the supervised loss on the labeled images, \ie, the cross-entropy loss,  and the $\mathcal{L}_{unsup}$ is the unsupervised loss for unlabeled images. $\lambda_{u}$ controls the weight of unsupervised term. 

\begin{figure*}
\color{black}
	\centering
  	 	\includegraphics[width=0.9\textwidth]{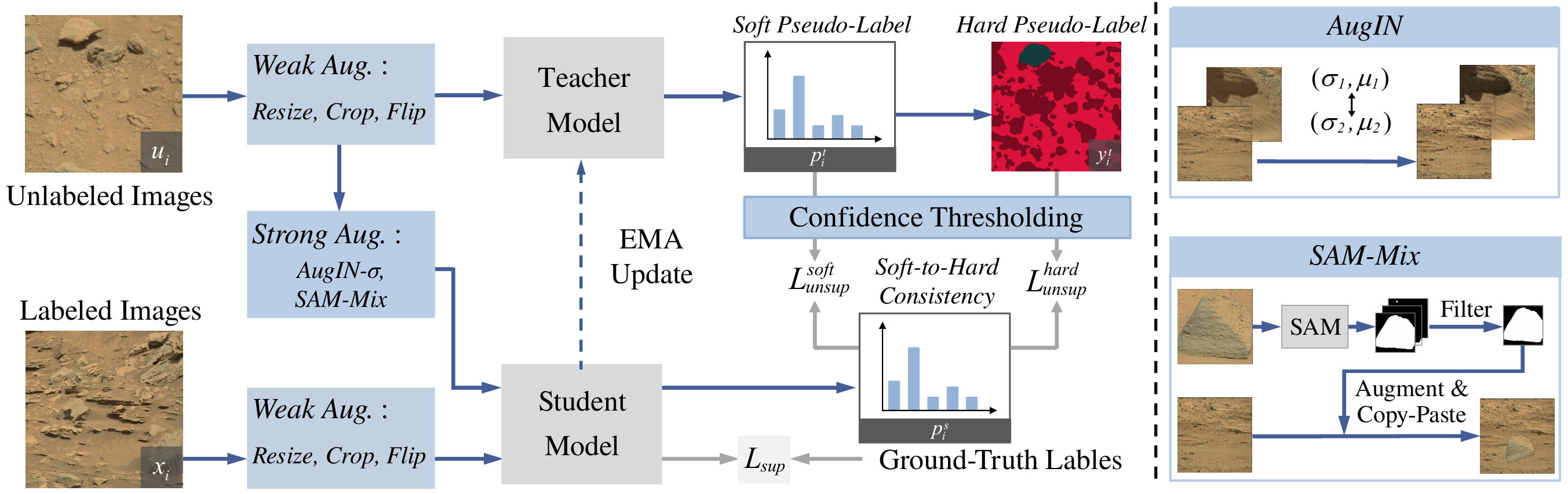}
	\caption{\revise{The overview of the proposed framework for semi-supervised Mars semantic segmentation.} We adopt a two-branch teacher-student architecture. Two novel augmentations are proposed as strong augmentations, \textit{AugIN} and \textit{SAM-Mix}. \textit{AugIN} exchanges the statistics of the two samples, \ie, mean and standard deviation. \textit{SAM-Mix} utilizes an off-the-shelf SAM to obtain the object binary masks to perform copy-paste operation, reducing the uncertainty of the augmented images. Finally, the model is optimized according to a soft-to-hard consistency learning strategy, utlizing both the soft labels $\textbf{p}_i^t$ and the hard labels $\textbf{y}_i^t$ based on the confidence.}
	\label{fig:arch}
\end{figure*}

Our method is based on the recent popular consistency regularization-based SSL method, FixMatch~\cite{sohn2020fixmatch}. Specifically, a two-branch network is adopted, consisting of a teacher model $f(\cdot;\theta_{t})$ and a student model $f(\cdot;\theta_{s})$. The teacher model $f(\cdot;\theta_{t})$ can be identical to the student model sharing the same weights. Alternatively, it can be updated gradually via the exponential moving
averaging (EMA) of the student model weights:
\begin{align}
    \theta_t \leftarrow m \theta_t + (1 - m) \theta_s,
\end{align}
where $m \in \left[ 0, 1 \right)$ is the momentum coefficient. We follow the EMA setting to update the teacher model, which is also recommended in Mean-Teacher~\cite{antti2017mean}. The student model is optimized via the backward gradients.

The core implementation in FixMatch is the weak-to-strong augmentation strategy, which serves as the perturbations and generate different augmented data views. Specifically, given the weak augmentations $\mathcal{T}_w$ and strong augmentations $\mathcal{T}_s$, the augmented views $\textbf{u}^{w}_i = \mathcal{T}_w(\textbf{u}_i)$, $\textbf{u}^s_i = \mathcal{T}_s(\textbf{u}_i)$ are constructed and fed into the teacher and student model to encode, respectively. The teacher model assigns the pseudo-labels for weakly augmented images, which are then utilized in the learning of student model for strongly augmented images. Concretely, the unsupervised consistency loss can be formulated as follows:
\begin{align}
    \label{eq:unsup_onehot}
    \mathcal{L}_{unsup}^{ce} &= \frac{1}{|\mathcal{B}_u|}\!\sum_{i=1}^{|\mathcal{B}_u|}\!\frac{1}{H\times W}\sum_{j=1}^{H\!\times\!W}\mathcal{L}_{ce}(\textbf{p}_i^s(j), \textbf{y}^t_i(j)),\\
    \textbf{y}^t_i(j) &= \mathbbm{1}(argmax(\textbf{p}^t_i(j))),
\end{align}
where $\textbf{p}_i^s(j)$/$\textbf{p}_i^t(j)$ is the predicted scores output by the student/teacher model after softmax layer corresponding to the $j_{th}$ pixel of the $i_{th}$ unlabeled image $\textbf{u}_i$. $\textbf{y}^t_i(j)$ is one-hot encoding of the pseudo-label generated from the teacher model and $\mathbbm{1}$ is the one-hot indicator function. $H$ and $W$ are the height and width of the image. $\mathcal{L}_{ce}$ is the cross-entropy loss function.

\textbf{2) Motivation:} For SSL in Mars image semantic segmentation, there are two main challenges to be solved: (a) Previous augmentations for the natural images on Earth can be ineffective due to the different properties of Mars images. (b) The unlabeled regions of the Mars images tend to be with high uncertainty, making the pseudo-labels less reliable for training. These problems affect the performance of the existing SSL frameworks for Mars image segmentation. 
To overcome these challenges, we propose a simple yet effective SSL framework, as shown in Fig.~\ref{fig:arch}, which adopts effective augmentations and learns semantic representations by exploring soft-to-hard consistency, which will be introduced in the following parts.

\subsection{Augmentations for Mars Images}\label{sec:aug}
As pointed in previous works~\cite{sohn2020fixmatch,zhao2023augmentation,yang2023revisiting}, the augmentation module plays an important role in SSL, encouraging the model to learn the consistency in the perturbations. Generally, the common augmentations adopted for SSL methods can be divided into the following categories:
\begin{itemize}
    \item \textbf{Geometrical Augmentation.} It utilizes some geometrical transformations, \eg \textit{Flip} and \textit{Translate}, to generate new data views. These augmentations often serve as the basic augmentations, \ie, the weak augmentations, due to their efficiency and stability.
    \item \textbf{Noise-Based Augmentation.} Different augmented views can be obtained by simply injecting random noise into the original image, \eg, \textit{Gaussian Noise}, and \textit{Random Mask}. 
    \item  \textbf{Color-Based Augmentation.} A series of color transformations are introduced to further enlarge the data distributions, \eg, \textit{Gaussian Blur}, \textit{Equalize}, and \textit{Sharpness}. More details can be found in~\cite{cubuk2020randaugment}. These transformations facilitate the model to learn the intrinsic semantic consistency by perturbing the color distribution of images.
    \item \textbf{Mixing-Based Augmentation.} Mixing methods have been proven effective for SSL scenarios. They mix the two samples via the interpolation (\textit{Mixup}~\cite{MixUp}) or cut-paste (\textit{CutMix}~\cite{CutMix}) operations. Some advanced mixing methods are further developed for SSL such as \textit{CowMix}~\cite{french2020milking} and \textit{ClassMix}~\cite{olsson2021classmix}.
    \item \textbf{Feature-Level Augmentation.} The most common augmentation in feature-level is \textit{Dropout}~\cite{srivastava2014dropout} operation, which can also be regarded as a kind of model perturbation. It is often utilized as strong augmentations in conjunction with other augmentations.
\end{itemize}

\begin{figure}
\color{black}
	\centering
  	 	\includegraphics[width=0.42\textwidth]{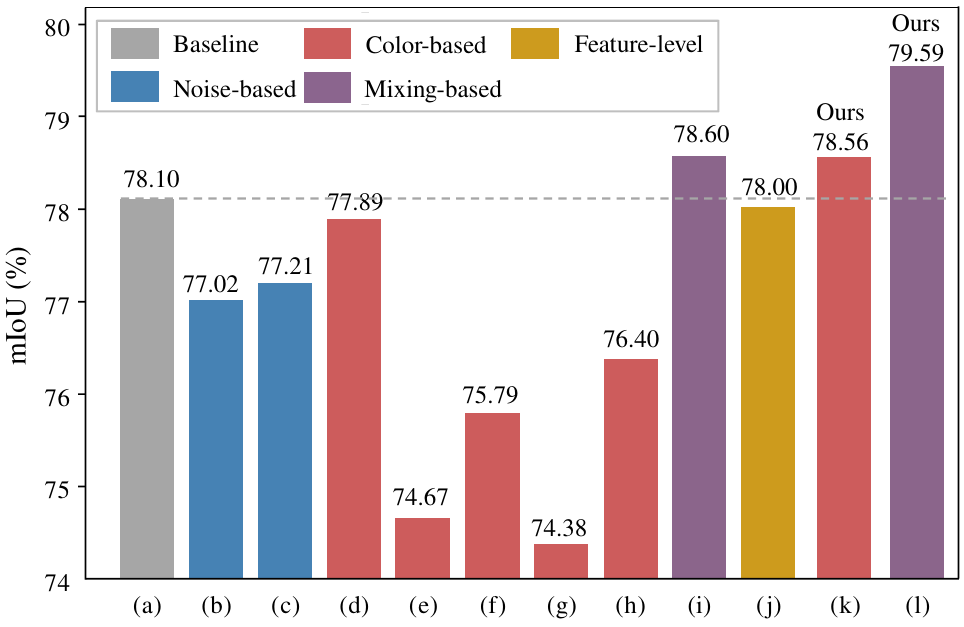}
	\caption{\revise{Comparison of different augmentations on SSL for Mars segmentation.} (a): Identity, (b): \textit{Gaussian Noise}, (c): \textit{CutOut}, (d): \textit{Gaussian Blur}, (e): \textit{Hue}, (f): \textit{Contrast}, (g): \textit{Equalize}, (h): \textit{Brightness}, (i): \textit{{CutMix}}, (j): \textit{Dropout}. (k) and (l) are the proposed \textit{AugIN} and \textit{SAM-Mix}.}
	\label{fig:aug_comp}
\end{figure}

\begin{table}[t]
  \centering
  \caption{Comparison of the statistical information between the Mars images and Earth images. Lower values of the metrics indicate less dispersion of the data distribution.
 } 
  \label{table:statstic}
\setlength{\tabcolsep}{3mm}{
    \begin{tabular}{c|c|c}
        \toprule
        \multirow{2.5}{*}{Dataset}&
\multicolumn{2}{c}{(R, G, B)}\\
\cmidrule(lr){2-3}
        & Standard Deviation & Variable Coefficient \\
        \midrule
          S$^{5}$Mars &(0.134, 0.121, 0.099) & (0.214, 0.233, 0.273)\\
        \midrule
          ImageNet~\cite{ImageNet} &(0.229, 0.224, 0.225) & (0.472, 0.491, 0.554)\\
        \bottomrule
    \end{tabular}
    }
\end{table}

\begin{figure}
	\centering
  	 	\includegraphics[width=0.5\textwidth]{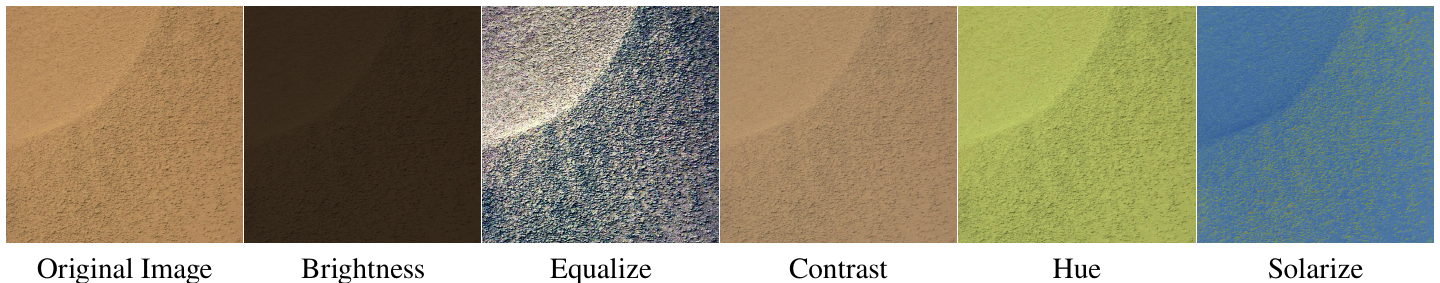}
	\caption{Examples of color-based augmented images. More details of the augmentations can be found in~\cite{cubuk2020randaugment}.}
	\label{fig:color_aug}
\end{figure}

We focus on the latter four, which serve as strong augmentations and have a significant impact on model performance. Following the recent work~\cite{zhao2023augmentation}, we adopt \textit{Resize}, \textit{Crop}, and \textit{Flip} as the weak augmentations. In addition, we choose different augmentations, which are commonly used and found beneficial for the learning of Earth images, as strong augmentations to demonstrate their impact separately. The results are shown in Fig.~\ref{fig:aug_comp}. As we can see, unlike natural images on Earth, the noise-, color-based and feature augmentations cannot bring a boost compared with the ``identity" baseline.
To further understand this phenomenon, we analyze the data from a statistical perspective and present the comparison of standard deviation and coefficient of variation between Mars and Earth images. As shown in Table~\ref{table:statstic}, the dispersion of RGB values in the Mars image is much less than that of Earth natural image, which indicates that the color distribution of the Mars image is more concentrated. It is in line with our observation that there is a high similarity within and between Mars images. \revise{Based on this conclusion, we argue that the traditional color-based perturbations lead to the color distribution shift of Mars images, causing the over-distortion problem~\cite{yuan2021simple} as shown in Fig.~\ref{fig:color_aug}, which is not conducive to the model segmentation learning.} Note that this is not trivial in the context of SSL because most previous SSL works adopt the color augmentations as a strong technique by default and lack specific consideration on the Mars images. Meanwhile, we empirically find that the feature perturbation \textit{Dropout} also fails to improve the performance, \revise{because it does not generate new input samples and cannot help the model learn richer semantic information. Besides, due to the irregular objects with occlusions and unclear contours, the model can face more serious uncertainty and consistency learning difficulty under the noise-based and random mixing-based augmentations, which will be discussed in the following.}

To this end, we propose two effective augmentations designed for Mars images, \textit{AugIN} and \textit{SAM-Mix}, and employ them in our method to boost the SSL performance.

\textbf{1) AugIN.}
To avoid drastic changes in image color distribution caused by direct perturbation, we propose \textit{AugIN} (Augment Instance Normalization), which generates augmented data views by exchanging statistics of different images, \ie, the mean and standard deviation. This is inspired by the successful practice of style transfer~\cite{huang2017arbitrary}. 
Specifically, given a image $\textbf{u}_i$ and a randomly sampled image $\textbf{u}_j$, we exchange the mean and standard deviation as follows:
\begin{equation}
AugIN(\textbf{u}_i, \textbf{u}_j)= \sigma\left(\textbf{u}_{j}\right)\left(\frac{\textbf{u}_{i}-\mu(\textbf{u}_{i})}{\sigma(\textbf{u}_{i})}\right) + \mu(\textbf{u}_{j}),
\end{equation}
where the $\mu(\cdot)$ and $\sigma(\cdot)$ are the mean and standard deviation functions. Meanwhile, we can spontaneously obtain the two variants, \textit{AugIN}-$\mu$ and \textit{AugIN}-$\sigma$, which only exchange the mean or standard deviation between two samples. In the implementation, we exchange the image statistics within the same batch following a randomly generated permutation. Note that the operation in our method that exchanges the statistics of images within the same batch does not change the statistics of the entire batch, which can be theoretically verified easily. This stabilizes the color distribution after augmentation and generates more reasonable augmented data. In contrast, traditional color augmentations change the statics directly without considering the whole color distribution, making the model suffer from the potential color distribution shift problem.

\begin{figure}
\color{black}
	\centering
  	 	\includegraphics[width=0.5\textwidth]{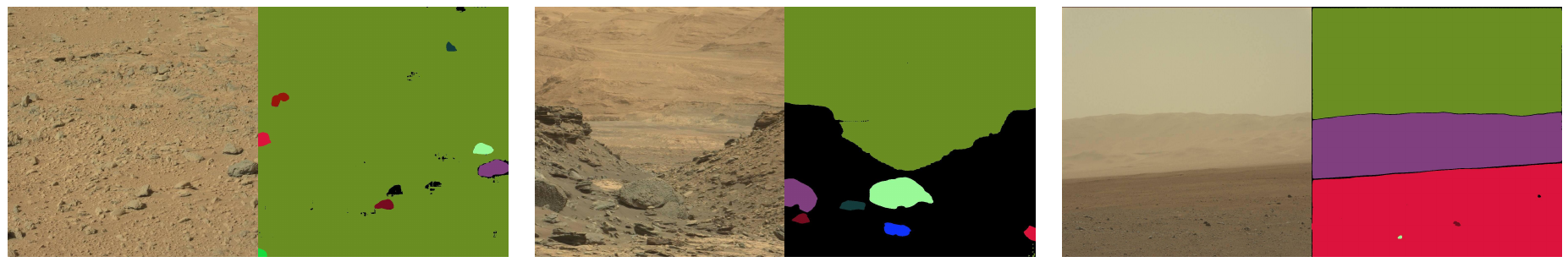}
	\caption{\revise{Examples of image-mask pairs. We show the filtered masks of each image with high predicted confidence. Note these masks are output by SAM in an instance-wise manner, and we illustrate them in different colors.}}
	\label{fig:sam}
\end{figure}

\textbf{2) SAM-Mix.} As shown in Fig.~\ref{fig:aug_comp}, \textit{CutMix} achieves a modest performance gain over the baseline, failing to meet the expected level of improvement. This is because there are many fragmentary objects with unclear edges in Mars images, and random cut-pasting manner may lead to high uncertainty, limiting the model performance. To this end, we propose \textit{SAM-Mix}, which is formulated as a generalization
of \textit{CutMix} using binary masks output by an off-the-shelf Segment-Anything Model (SAM)~\cite{kirillov2023segment}. 

SAM attracts lots of attention recently,
which can produce binary masks for the objects in an image from input or randomly generated prompts. We utilize an off-the-shelf SAM to produce a mask of the target object and paste it into the source image. Compared with random rectangular mask generation, \revise{SAM can generate high-quality masks to segment specific objects as shown in Fig.~\ref{fig:sam}. Specifically, given an image, a list of binary masks with the corresponding confidence score is output by SAM. These masks are first filtered so that 1) the size of the mask is limited to a certain range and 2) the confidence of the mask is above a certain threshold. If there is no qualified mask, a random rectangular mask will be directly generated. Then a Gaussian filter is applied to the masks to eliminate possible noise. Subsequently, we randomly select a qualified mask and further transform the masked object, \ie, \textit{Rotation}, \textit{Flip}, and \textit{Rescaling}. The pasting position will not be adjusted, that is, it will generally pasted corresponding to the position of the original image, to avoid some unreasonable cases, \eg, the sky appearing in the bottom half of the image. The corresponding segmentation labels are also generated in the same way, which are used to train the model as previous work~\cite{french2019semi}.
}

We note that \textit{SAM-Mix} shares similarities with other segmentation-based mixing augmentation strategies~\cite{olsson2021classmix,dwibedi2017cut,tripathi2019learning,fang2019instaboost,yang2020fda}, which develop the binary mask generation in an instance-wise or class-wise manner. However, in contrast to the above mixing methods, \textit{SAM-Mix} gets rid of the reliance on ground-truth labels, making it possible for the augmentation on unlabeled images. Furthermore, SAM's strong generalization ability enables us to produce high-quality masks for individual objects efficiently, which is compatible with images of Mars that contain multiple objects simultaneously. \textit{SAM-Mix} reduces the uncertainty caused by random mixing and further improves the performance of the model.

\subsection{Soft-to-Hard Consistency Learning}
\label{sec:sth}
As mentioned in the previous section, the Mars images are with more confusing categories, such as sand and soil, rock and bedrock, which require a more fine-grained representation learning target, especially for the unlabeled regions with high uncertainty in our dataset. Meanwhile, for the data collection and annotation, it is more difficult to obtain large-scale and high-quality annotated Mars images than natural Earth images, due to the complexity of the Mars terrain, the required expert knowledge, and the limited transmission bandwidth. Therefore, previous works using only unlabeled regions with high confidence for training can be sub-optimal in the Mars SSL context.

To this end, we propose a soft-to-hard consistency learning strategy, which utilizes both the soft and hard pseudo-labels according to a confidence thresholding policy.
\color{black}The hard pseudo-label is the one-hot label representation $\textbf{y}_i^t(j)$ in Eq.~\ref{eq:unsup_onehot}, which is obtained by the $argmax(\cdot)$ operation. The soft label is represented as the model prediction scores $\textbf{p}_i^s(j)$, which denotes the probability distribution over different semantic categories. 
Specifically, the optimization objective for the soft pseudo-label can be formulated as:
\begin{align}\label{eq:relation2}
\color{black}
    \mathcal{L}_{unsup}^{soft} = -\frac{1}{|\mathcal{B}_u|}\!\sum_{i=1}^{|\mathcal{B}_u|}\!\frac{1}{H\times W}\sum_{j=1}^{H\!\times\!W} \textbf{p}_i^t(j)\log (\textbf{p}_i^s(j) ).
\end{align}
\revise{Intuitively, Eq.~\ref{eq:relation2} optimizes the similarity of the two distributions, \ie, ${p}_i^t$ and ${p}_i^u$, which indicate the predicted class probability of the teacher and student model.} Based on this objective, we can further find that:

\noindent$\bullet$ When $max(\textbf{p}_i^t(j)) \approx 1$, the teacher model assigns the pseudo-labels with high confidence, and the Eq.~\ref{eq:relation2} degenerates to be almost equivalent to Eq.~\ref{eq:unsup_onehot}.

\noindent$\bullet$ When $max(\textbf{p}_i^t(j)) < 1-t$ where $t$ is a positive constant, the predictions of the teacher model is less confident. This objective encourages the student model to learn the consistency measured by the relevance of current features to different prototype anchors. This can be seen as a more fine-grained smooth label of the unknown regions in Mars images, which can belong to a new class or the old class with high uncertainty.

Therefore, the hard label provides a confident target to force the model to predict a distribution with low entropy, learning the explicit semantic mapping in images.
In contrast, the soft label objective encourages the model to learn the consistency in a more gentle way, which can be viewed as performing the self-distillation~\cite{park2019relational} of relational knowledge, modeled as the feature similarity to the prototype features stored in the weights of the classification head. This allows the model to make better use of unlabeled data to improve the representation consistency learning in an unsupervised manner, achieving a better representation space. 

Based on the above analysis, we propose a confidence-based thresholding policy to integrate the two objective functions organically. We utilize the hard pseudo-labels in high-confidence regions while using soft pseudo-labels in low confidence regions, fully taking advantage of the training signals from unlabeled data. Specifically, we first obtain the confidence score of teacher model predictions as $max(\textbf{p}_i^t(j))$. Then, the student model is optimized as follows ($t_{hard}$ and $t_{soft}$ are the threshold hyper-parameters):

\vspace{0.2em}
\noindent 1) If $max(\textbf{p}_i^t(j)) > t_{hard}$, Eq.~\ref{eq:unsup_onehot} is applied to optimize the model with the highly confident one-hot pseudo-label; 

\noindent 2) If $max(\textbf{p}_i^t(j)) < t_{soft}$, the soft label objective is optimized, to avoid noisy signals from other prototype features in the high confidence region.
\vspace{0.2em}

Finally, the model is optimized in an end-to-end manner using the objective in Eq.~\ref{equ:loss:total}. The supervised term $\mathcal{L}_{sup}$ is the cross-entropy loss on the labeled images. The whole consistency regularization term $\mathcal{L}_{unsup}$ is:
\begin{align}
    \mathcal{L}_{unsup} = \mathcal{L}_{unsup}^{hard} + \lambda_{s} \mathcal{L}_{unsup}^{soft},
\end{align}
where $\mathcal{L}_{unsup}^{hard}$ is exactly the $\mathcal{L}_{unsup}^{ce}$ in Eq.~\ref{eq:unsup_onehot} and $\lambda_{s}$ is the weight coefficient.

\section{\newtxt{Experiments and Results}}
\label{sec:exp}

\subsection{Dataset}
We use the proposed S$^{5}$Mars dataset and AI4Mars dataset, which are introduced in the Section~\ref{sec:dataset}. For semi-supervised learning evaluation, we adopt a stratified sampling strategy to extract different proportions of data from the dataset as labeled data and the rest as unlabeled data, to jointly train our model. 
Note that all methods are evaluated under the same data partition lists.

\subsection{Implementation Details and Metrics}
\zjh{Our model is based on DeepLabV3+~\cite{DeeplabV3_plus}, adopting a ResNet-50~\cite{he2016deep} pre-trained on Image-Net~\cite{ImageNet} as the segmentation backbone. 
We use an output stride of 16 by default.
The batch size is set to 8. An SGD optimizer with a momentum of 0.9. A polynomial learning-rate decay with an initial value of 0.01 are adopted to train the student model. Specifically, the learning rate is scaled by $(1-{iter}/{max\_iter})^{0.9}$. The EMA momentum coefficient $m$ is set as $min(1-{1}/{(iter+1)}, 0.996)$ following the~\cite{zhao2023augmentation}.
$\lambda_{r}$ and $\lambda_{unsup}$ are set to 1.0 and 2.0 by default. The model is trained for 240 epochs by default and the teacher model is used for the evaluation.
The images for training are cropped to the size of 512$\times$512. The test images are center-cropped to 1024$\times$1024 size.
We train our model on a single NVIDIA RTX 3090 GPU.}

{We evaluate performance using \textbf{Mean Pixel Accuracy} (mAcc) and \textbf{Mean Intersection over Union} (mIoU) as the metrics.
}

\begin{table*}[t]
\centering
\color{black}
\taburulecolor{black}
{\caption{Segmentation performance on the S$^{5}$Mars using the ResNet-50 as the backbone.}}
\label{table:semi_pku}
\setlength{\tabcolsep}{3.0mm}{
\begin{tabular}{c|c|c|c|c|c|c}
\toprule
\multirow{2}{*}{Method}&
\multicolumn{2}{c|}{20\% data}&
\multicolumn{2}{c|}{50\% data} & \multicolumn{2}{c}{100\% data}\\
\cmidrule(lr){2-3}\cmidrule(lr){4-5}\cmidrule(lr){6-7}
   & mAcc (\%) & \makecell[c]{mIoU (\%)} & \makecell[c]{mAcc (\%)} & \makecell[c]{mIoU (\%)} &\makecell[c]{mAcc (\%)} & \makecell[c]{mIoU (\%)}  \\
\midrule

    \revise{Supervised} &76.71& 66.26& 80.17& 72.65&	82.12 & 75.04 \\
\midrule
    MT~\cite{antti2017mean} & 76.89 & 70.92	& 81.95 &75.93 &	82.32 &77.87 \\
    MT~\cite{antti2017mean} + ClassMix~\cite{olsson2021classmix} & 77.38 & 71.54 & 82.34 &76.42 &83.21 &78.04 \\
    FixMatch~\cite{sohn2020fixmatch} &77.80 & 71.08	& 79.47& 72.98 & 80.75 &73.69 \\
    RanPaste~\cite{wang2021ranpaste} &76.36&66.66&80.51&74.08&	82.28 & 75.78\\
    U$\rm ^2$PL~\cite{wang2022semi}&78.03&72.32&	82.29&77.56&83.41&78.60\\
    AugSeg~\cite{zhao2023augmentation}& 78.28& 72.38& 81.52 & 75.40 &	81.82 &76.90 \\
    UniMatch~\cite{yang2023revisiting}& 76.92& 69.83 & 78.53&71.56 &	79.82& 72.60 \\
    
\midrule
    \textbf{Ours} &\textbf{82.85} &\textbf{76.49} &\textbf{83.56} &\textbf{78.66}	&\textbf{84.73}&\textbf{80.15} \\
    
\bottomrule
\end{tabular}
}
\end{table*}

\begin{table}[t]
\color{black}
\centering
\caption{Segmentation performance on the AI4Mars.}
\label{table:semi_ai4mars}
\begin{tabular}{c|c|c|c|c}
\toprule
\multirow{2.5}{*}{Method}&
\multicolumn{2}{c|}{20\% data}& \multicolumn{2}{c}{100\% data}\\
\cmidrule(lr){2-3}\cmidrule(lr){4-5}
   & mAcc & \makecell[c]{mIoU} & \makecell[c]{mAcc} & \makecell[c]{mIoU} \\
\midrule
    Supervised &71.68&66.14&	74.43&68.34\\
\midrule
    \revise{MT}~\cite{antti2017mean} &  75.44 & 70.13 &77.82 &71.98 \\
MT~\cite{antti2017mean}+ClassMix~\cite{olsson2021classmix} & 76.86&70.49&	78.56 &72.67\\
    FixMatch~\cite{sohn2020fixmatch} &76.27&70.36&	77.37&71.90\\
    RanPaste~\cite{wang2021ranpaste}&75.16 &70.37 & 77.39&70.59 \\
    \revise{U$\rm ^2$PL}~\cite{wang2022semi}& 77.11 & 70.89 &78.62 & 72.41\\
    AugSeg~\cite{zhao2023augmentation}& 76.88 & 70.15 &	77.45 &72.34 \\
    \revise{UniMatch}~\cite{yang2023revisiting}&75.60 &70.24 & 77.21& 71.36\\
\midrule
    \textbf{Ours} &\textbf{77.60} & \textbf{71.79}& \textbf{80.33}& \textbf{74.68} \\
\bottomrule
\end{tabular}

\end{table}

\begin{table*}[t]
\color{black}
\centering
\caption{\revise{Segmentation performance of different classes on the S$^{5}$Mars using all labeled training data.}}
\label{table:class_iou}
\begin{tabular}{c|ccccccccc|c}
\toprule
\multirow{2.5}{*}{Method}&
\multicolumn{9}{c|}{Class IoU (\%)}&\multirow{2.5}{*}{mIoU}\\
\cmidrule(lr){2-10}
   &Sky&Ridge&Soil	&Sand&Bedrock&Rock&Rover&Trace&Hole&\\
\midrule
    FixMatch~\cite{sohn2020fixmatch} &88.97&90.06&83.05&74.19&92.09&20.31&52.54&88.75&73.26&73.69\\
    U$\rm ^2$PL~\cite{wang2022semi} &94.86&93.25&85.80&81.02&92.18&23.55&93.77&78.06&64.96&78.60\\
    AugSeg~\cite{zhao2023augmentation}& 94.75	&93.00&84.78&77.36&92.02&10.33&83.08&81.00&75.84&76.90\\
    Ours &95.64	&94.20	&87.18	&83.18&92.42&22.89&87.17&	85.60&74.10 &\textbf{80.15}\\
\bottomrule
\end{tabular}
\end{table*}

\begin{table}[t]
  \centering
  \caption{Comparison with the zero-shot models on S$^{5}$Mars Dataset.} 
  \label{table:zero-shot}
\setlength{\tabcolsep}{2.0mm}{
    \begin{tabular}{c|ccc}
        \toprule
        Methods  & mAcc (\%) &  mIoU (\%) 
        &Inference Time\\
        \midrule
        SAM-CLS~\cite{kirillov2023segment} &36.38 &28.53 &691 ms/img\\
        \midrule
        SAN~\cite{xu2023side} &14.70 & 3.28 & 102 ms/img
        \\
        SAN-FT~\cite{xu2023side} &11.60 & 10.54 &102 ms/img \\
         \midrule
          Ours &\textbf{84.73}&\textbf{80.15} & 19 ms/img\\
        \bottomrule
    \end{tabular}
    }
\end{table}

\subsection{Comparison Results}
\textbf{Compared with the SSL Methods}.
\zjh{We compare our model with state-of-the-art semi-supervised learning methods, Mean Teacher~\cite{antti2017mean} (MT), ClassMix~\cite{olsson2021classmix}, FixMatch~\cite{sohn2020fixmatch}, RanPaste~\cite{wang2021ranpaste}, U$\rm ^{2}$PL~\cite{wang2022semi}, 
AugSeg~\cite{zhao2023augmentation}, and UniMatch~\cite{yang2023revisiting}, covering the latest consistency regularization-based and contrastive learning-based methods as well as the naive supervised training results, which only utilize the labeled data. We conduct these compared methods with their official implementations (except for the MT which we adopt a better training hyper-parameters for fair), using the same backbone ResNet-50. As shown in Table~\ref{table:semi_pku} and Table~\ref{table:semi_ai4mars}, our method achieves the best performance across different labeled data ratios and datasets. MT~\cite{antti2017mean} uses \textit{Gaussian noise} and \textit{Dropout} as augmentations for both teacher and student branches. However, earlier works do not use weak-to-strong augmentation strategy, which makes them sub-optimal.
We also employ the ClassMix~\cite{olsson2021classmix} augmentation with MT. But the quality of the generated mixed image strongly depends on the quality of the pseudo-labels, which cannot be guaranteed on the Mars images with unclear object contours.
FixMatch~\cite{sohn2020fixmatch} and UniMatch~\cite{yang2023revisiting} perform even worse than the supervised baseline when the more labeled data are available. This is mainly because they employ a shared encoder instead of the teacher-student architecture, which we find is less effective in our setting. Besides, AugSeg~\cite{zhao2023augmentation} achieves state-of-the-art performance on Earth image benchmarks, while it is not satisfying in Mars semantic segmentation task due to the adopted various color augmentations. As for the contrastive learning-based methods, U$^{2}$PL uses filtered pseudo-labels to perform the pixel-wise contrastive learning. However, the training cost of such methods is generally high, which will be discussed in Section~\ref{sec:complex}. 
}

\revise{We also present the segmentation performance of each class in Table~\ref{table:class_iou}. Note that the optimal data augmentations and learning strategy may differ across different categories. Remarkably, our method can achieves the best results on the head classes. Meanwhile, the performance on the tail classes, of which the sample number is small, is also comparable with other methods. }

\textbf{Compared with Zero-Shot General Models.} Recently, the general large models for segmentation achieves a great success, which can deal with the unseen data in training with the help of massive amounts of training data or the help of the vision-language model, \eg, CLIP~\cite{radford2021learning}. Here we highlight that current zero-shot learning methods or open-vocabulary methods are still unable to handle the Mars image semantic segmentation task well due to the fine-grained feature classification and required expert knowledge. 

SAM~\cite{kirillov2023segment} can produce the high-quality object binary masks. To obtain the corresponding label, we apply a classification head subsequent to the image encoder. Specifically, the pre-trained image encoder in SAM is fixed and we train a classification decoder, which takes the encoded feature as input and outputs the pixel-wise semantic label, denoted by SAM-CLS. We can find that the extracted features by the encoder are not discriminative for Mars semantic segmentation, as shown in Table~\ref{table:zero-shot}.
As for vision-language models, we compare with state-of-the-art method, SAN~\cite{xu2023side}. We evaluate the model with both official model weights and fine-tuned weights on the target dataset, denoted as SAN-FT. However, poor performance is obseverd as shown in Table~\ref{table:zero-shot}. This is mainly due to two aspects: 1) the Martian terrain category is relatively rare in the corpus, 2) domain-specific expert knowledge is required for the fine-grained classification, which lead to the difficulty of feature space alignment for the Mars segmentation dataset.
Moreover, their high training and inference overhead makes them sub-optimal for resource-constrained extraterrestrial tasks.

Overall, our method achieves remarkable performance, which verifies the effectiveness of the proposed method.

\begin{table}[t]
  \color{black}
            \vspace{0pt}
		\parbox{.46\linewidth}{
            \vspace{0pt}
		\caption{Ablation studies on \textit{AugIN}. Identity denotes the method without \textit{AugIN}.}
  \label{tab:augin}
  \centering
\setlength{\tabcolsep}{2mm}{
  \begin{tabular}{c|cc}
  \toprule
  Method & mAcc& mIoU \\
  \midrule
   Identity &84.20 &79.59\\
   \revise{EFDM}~\cite{zhang2022exact}& 83.97 & 78.78\\
   \revise{WCT}~\cite{li2017universal} & 83.38 &78.83\\
   \revise{FDA}~\cite{yang2020fda} & 84.20 & 78.70 \\
   
   \midrule
   \textit{AugIN}-$\mu$ &83.64 &78.88 \\
   \textit{AugIN}-$\sigma$ & \textbf{84.73}&\textbf{80.15}\\ 
  \textit{AugIN}& 83.58 & 78.69\\
  \bottomrule
  \end{tabular}}
}
  \label{tab:hard_t}
		\hfill
            \vspace{0pt}
		\parbox{.51\linewidth}{
			\caption{Ablation studies on different mixing methods. Identity denotes the method without any mixing augmentation.}
  \label{tab:mix}
  \centering
\setlength{\tabcolsep}{2mm}{
  \begin{tabular}[t]{c|cc}
  \toprule
  Method & mAcc& mIoU\\
  \midrule
   Identity &83.33 & 78.56\\
   \midrule
   \textit{CutMix~\cite{CutMix}}&83.85 & 78.98 \\
   \textit{ClassMix~\cite{olsson2021classmix}}  &83.81 & 78.89\\ 
   \revise{\textit{DACS}}~\cite{tranheden2021dacs} &83.80 &78.69  \\
  \textit{SAM-Mix} (Ours)& \textbf{84.73} & \textbf{80.15}\\
  \bottomrule
  \end{tabular}}
  \label{tab:soft_t}
		}\hfill
	\end{table}

\begin{table}[t]
\centering
\begin{minipage}[!t]{1\columnwidth}
  \caption{Ablation studies on the object augmentations in \textit{SAM-Mix}.}
  \label{tab:obj_aug}
  \centering
\setlength{\tabcolsep}{3mm}{
  \begin{tabular}{ccc|cc}
  \toprule
  \textit{Flip}&\textit{Rotate} &\textit{Rescaling} & mAcc (\%) & mIoU (\%) \\
  \midrule
    & & & 84.57&79.46\\
   \Checkmark& & &\textbf{84.81} &79.71 \\
   \Checkmark &\Checkmark & & 84.80 &79.88\\ 
  \Checkmark&\Checkmark &\Checkmark & {84.73} & \textbf{80.15}\\
  \bottomrule
  \end{tabular}}
  \end{minipage}
\end{table}

\begin{figure}[t]
	\centering
  	 	\includegraphics[width=0.35\textwidth]{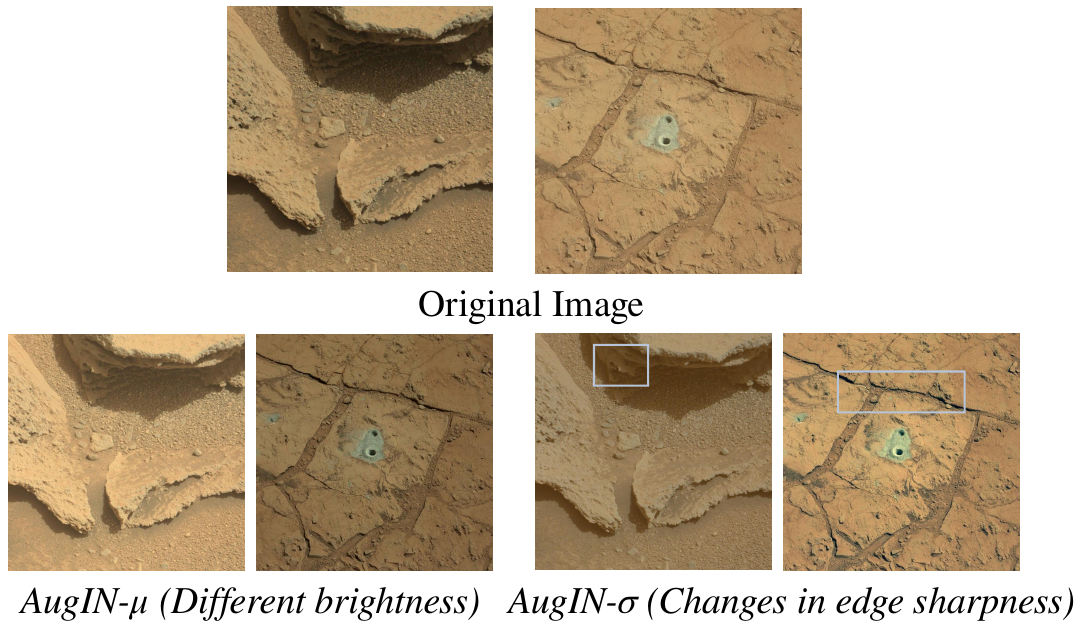}
	\caption{Examples of augmented results by \textit{AugIN}. \textit{AugIN}-$\mu$ mainly affects the image brightness while \textit{AugIN}-$\sigma$ changes the sharpness of the object edges.}
	\label{fig:augin}
\end{figure}

\begin{table}[t]
  \centering
  \caption{Ablation studies on soft and hard pseudo-labels.}
  \label{table:soft_hard}
\setlength{\tabcolsep}{2mm}{
    \begin{tabular}{cc|cc}
        \toprule
        Soft Label& Hard Label& mAcc (\%)& mIoU (\%) \\
        \midrule
         \Checkmark&  &83.36 & 78.12\\
          &\Checkmark&  84.28 & 79.70\\
           \Checkmark&\Checkmark& \textbf{84.73}&\textbf{80.15} \\
        \bottomrule
    \end{tabular}
    }
\end{table}

\begin{figure}[t]
	\centering
    \includegraphics[width=0.48\textwidth]{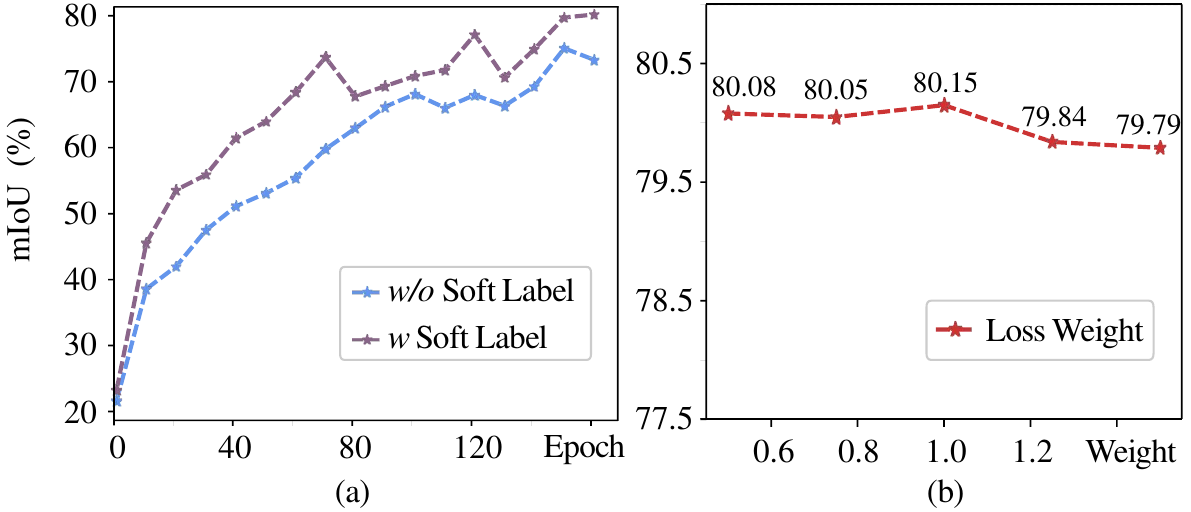}
	\caption{(a): The effect of soft pseudo-labels on performance of different epochs. Faster convergence is observed when equipped with soft labels. (b) Ablation study results on the loss weight of soft pseudo-labels.}
	\label{fig:epoch}
\end{figure}

\subsection{Ablation Studies}
\zjh{In the following, we conduct a series of ablations studies on the S$^{5}$Mars dataset using full data by default. ResNet-50 is adopted as the backbone.}

\zjh{\noindent\textbf{Effect of AugIN.} Recall that \textit{AugIN} exchanges the mean or standard deviation of different samples as shown in Fig.~\ref{fig:augin}. 
First, we give the ablation studies on the different variants of the \textit{AugIN} in Table~\ref{tab:augin}. \textit{AugIN}-$\mu$, \textit{AugIN}-$\sigma$ and \textit{AugIN} denote swapping the mean, standard deviation, and both, respectively. 
As we can see, \textit{AugIN}-$\sigma$ can bring boost to the model performance, while swapping mean shows adverse effect. By comparing the performance of different categories, we find that the main performance degradation comes from the hole, rover, and rock. 
We argue that this is because the mean of a image corresponds to the brightness. Exchanging the mean can cause the inappropriate brightness changes of objects. For example, the brightness of rover is obviously different from that of rock, and corrupting this information is not conducive to distinguish objects. In contrast, the standard deviation mainly affects the degree of dispersion of the data while maintaining the overall brightness, which is mainly reflected in the clarity of the object edge. This helps the model to produce better prediction results at the object edge. We finally choose \textit{AugIN}-$\sigma$ in implementation. 

\revise{Besides, to give a comprehensive evaluation, we provide the results of replacing \textit{AugIN} with other methods~\cite{zhang2022exact,li2017universal,yang2020fda} that exchange inter-image information. FDA~\cite{yang2020fda} exchanges low-frequency information which is similar to \textit{AugIN}-$\mu$. Other style-transfer-based methods, \ie, EFDM~\cite{zhang2022exact} and WCT~\cite{li2017universal}, also fail to improve the performance because they still cannot avoid the caused color distribution shift.} 
}

\zjh{\noindent\textbf{Effect of SAM-Mix.}
Table~\ref{tab:mix} gives the analysis of the \textit{SAM-Mix} augmentation. The main difference between these approaches lies in the way the masks are generated. However, these generated masks have high uncertainty, making it difficult for the model to learn consistency in SSL tasks. \textit{CutMix} randomly generates a rectangular mask from the beta distribution. \textit{ClassMix} takes the predicted region of a certain category as the mask through the generated pseudo-labels, \revise{while \textit{DACS} uses the ground-truth labels to mix images. However, these methods are performed in class-wise instead of instance-wise. For example, all rocks would be cut and pasted to another image, which can cause serious occlusions and increase the difficulty of consistency learning. Meanwhile, \textit{ClassMix}, which relies on pseudo-labels, still cannot provide good guidance in the early stage of training, while \textit{DACS}, which is based on ground-truth labels, heavily relies on the number of labels and can only generate limited mixed samples.} 
In contrast, we utilize the masks output by SAM to locate the objects and filter out the mask with higher confidence, achieving better performance. 

Meanwhile, we present the effect of the object-wise augmentations, \ie, \textit{Rotation}, \textit{Flip}, and \textit{Rescaling} for the cut out object. As shown in the Tabel~\ref{tab:obj_aug}, these simple available geometrical object augmentations can bring further improvement slightly by promoting the mask diversity.
}

\zjh{\noindent\textbf{Soft-to-Hard Pseudo-Label.} To depict the semantic features in a more fine-grained manner, we utilize the soft pseudo-labels in addition to the hard labels for the unlabeled images. 
Table~\ref{table:soft_hard} shows the effect of the soft and hard label, respectively. As we can see, the model gives the best results when both kinds of labels are utilized. We note that only employing the soft labels can not yield a good performance, which illustrates that the hard label is important for its low entropy constraints on the model output predictions. On the other hand, soft labels serve as a useful complement to hard labels, especially by making full use of the supervised signals in low-confidence regions. It promotes consistency learning by using the correlation between features and prototypes of different categories as the objective, which can be deemed as a relational knowledge distillation process.

Meanwhile, we note that the dual label optimization strategy can significantly improve the convergence speed of the model as shown in Fig.~\ref{fig:epoch} (a). This is because of the extra supervisory signal provided by those low-confidence regions, which would be discarded using the hard labels.
}

\begin{table}[t]
\color{black}
            \vspace{0pt}
		\parbox{.48\linewidth}{
            \vspace{0pt}
		\caption{Ablation study on the threshold $t_{hard}$.}
			\centering
			\setlength{\tabcolsep}{6pt}
   \setlength{\tabcolsep}{3mm}{
			\begin{tabular}{c|c}
   \toprule
    $t_{hard}$ & mIoU (\%)\\
    \midrule
    \revise{0.9} & 78.82\\
    0.7& 79.31\\
    0.6 & {79.87}\\
    0.5 & \textbf{80.15}\\
    0.4 & {80.00}\\
    \bottomrule
\end{tabular}
}
  \label{tab:hard_t}
		}\hfill
            \vspace{0pt}
		\parbox{.48\linewidth}{
			\caption{Ablation study on the threshold $t_{soft}$.}
			\centering
			\setlength{\tabcolsep}{6pt}
    \setlength{\tabcolsep}{3mm}{			
   \begin{tabular}{c|c}
   \toprule
    $t_{soft}$ & mIoU (\%)\\
    \midrule
    1.0& 79.83\\
    0.95 & 79.75\\
    0.9 & \textbf{80.15}\\
    0.8 & 80.04\\
    \revise{0.6} & 79.58\\
    \bottomrule
\end{tabular}
}
  \label{tab:soft_t}
		}\hfill
	\end{table}

\zjh{\noindent\textbf{Confidence-based Thresholding Strategy.} We utilize hard pseudo-labels in highly confident regions ($>t_{hard}$) while using soft labels in regions with low confidence ($< t_{soft}$). Table~\ref{tab:hard_t} gives the results of different thresholds. 
For the hard labels, a too-high threshold \revise{($>0.7$)} can significantly reduce the number of training labels, limiting the effect of pseudo-labels. \revise{Meanwhile, a too-low threshold can degrade the performance slightly, due to the additional introduced uncertain pseudo-labels in model consistency learning. However, this effect is relatively weak because only few pixels are involved.}
For the soft labels, we discard the high confidence region. Because these regions can be considered reliable, and the model should give predictions with high confidence. In this case, correlations generated with other semantic prototypes are often noise, which is not conducive to consistency learning as shown in Table~\ref{tab:soft_t}. \revise{However, a too-low threshold degrades the performance significantly, which is because these learning targets with high uncertainty lead to an unstable learning process.}
}

\zjh{\noindent\textbf{Loss Weight.} 
We further conduct the ablation studies on the loss weight $\lambda_s$ of the soft pseudo-labels $\mathcal{L}_{unsup}^{soft}$ in Fig.~\ref{fig:epoch} (b). As we can see, the model performance degrades when the loss weight is too large. This indicates the hard pseudo-label signals of the high-confidence regions are important to the model, which constrain the model predictions to be with low entropy. For the loss weight $\lambda_{unsup}$, we set it to be 2.0 following the previous works~\cite{zhao2023augmentation,sohn2020fixmatch} to make our method more general. 
}

\begin{table}[t]
  \centering
  \caption{Complexity Analysis on S$^{5}$Mars Dataset.} 
  \label{table:complexity}
\setlength{\tabcolsep}{2.3mm}{
    \begin{tabular}{c|ccc}
        \toprule
        Methods  & Training Time &  Params
        &mIoU (\%)\\
        \midrule
        U$\rm ^2$PL~\cite{wang2022semi}&98.7h & 196.15M & 78.60\\
        FixMatch~\cite{sohn2020fixmatch} &34.3h & 40.47M & 73.69\\
        AugSeg~\cite{zhao2023augmentation} &35.9h & 80.94M & 76.90\\
        Ours &42.8h & 80.94M & \textbf{80.15}\\ 
        \bottomrule
    \end{tabular}
    }
\end{table}

\begin{figure*}[t]
	\centering
  	 	\includegraphics[width=0.85\textwidth]{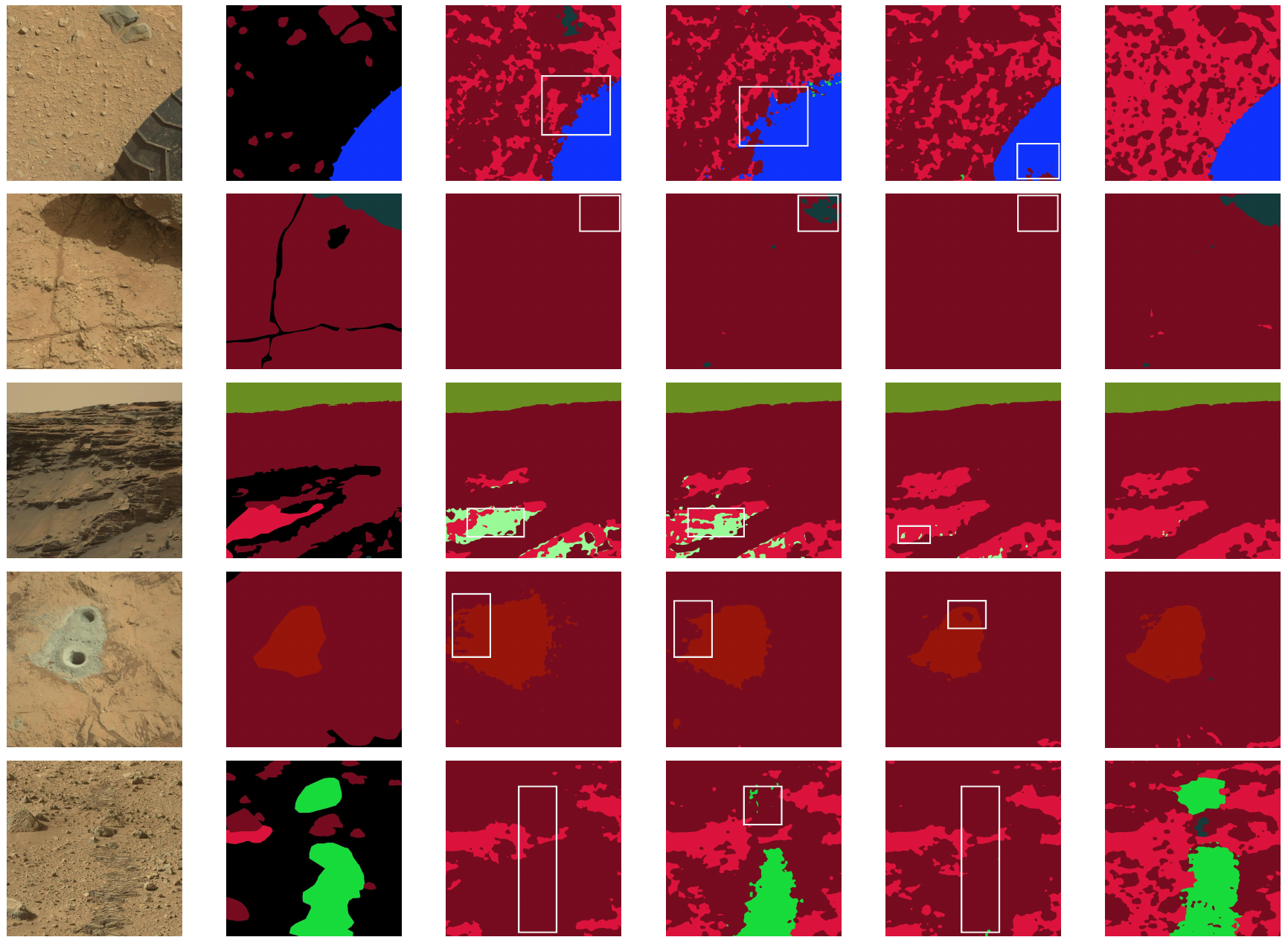}
	\caption{Qualitative results on S$^{5}$Mars dataset with full labels. Columns from left to right denote the original images, the ground-truth, the supervised results, the MT~\cite{antti2017mean}+ClassMix~\cite{olsson2021classmix} results, AugSeg~\cite{zhao2023augmentation} results, and our method results, respectively.}
	\label{fig:seg_compare}
\end{figure*}

\begin{figure}
\color{black}
	\centering
\includegraphics[width=0.4\textwidth]{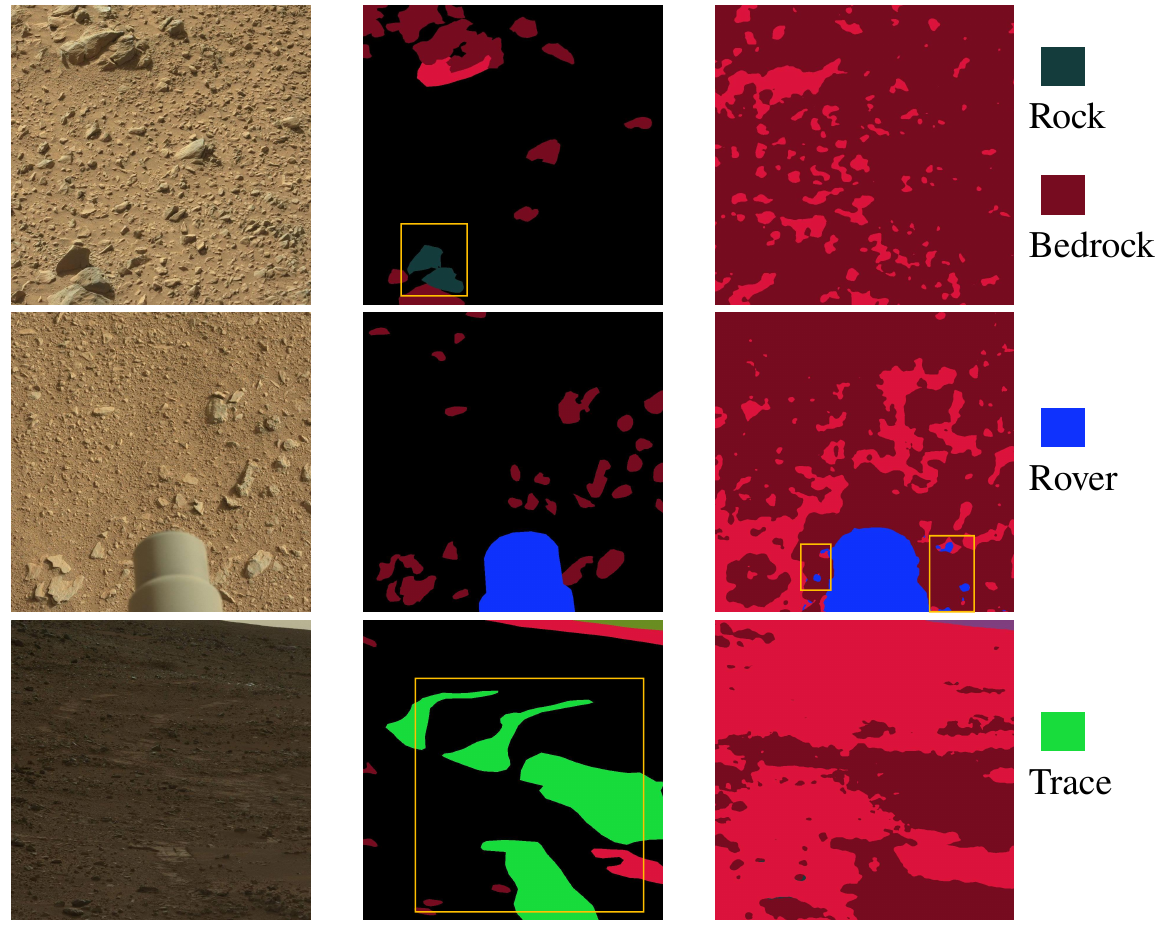}
	\caption{Some failure cases of our method. From left to right are the original image, the ground-truth label, and the predicted result.}
	\label{fig:failure}
\end{figure}

\noindent\textbf{Complexity Analysis.}\label{sec:complex}
We first present the results of U$\rm ^2$PL~\cite{wang2022semi} based on contrastive learning in Table~\ref{table:complexity}. These methods have high complexity because they need to maintain a large number of samples in the memory bank and perform instance discrimination tasks with multiple negative samples. As we can see, our method based on consistency regularization has an obvious complexity advantage, which shows the advantage of this type of method for extraterrestrial missions.
Compared with other consistency-regularization methods~\cite{sohn2020fixmatch,zhao2023augmentation}, our method has a longer training time due to the additional back-propagation process for soft pseudo-label optimization and the data augmentations. Since the same backbone network is used, the inference time is the same across these methods.
As for the model parameters, the difference lies in whether the teacher and student models share parameters.
Despite the cost incurred in terms of training time and space parameters, we argue that they are acceptable where a significant performance gain is observed.

\subsection{Qualitative Results}
We present subjective segmentation results in Fig.~\ref{fig:seg_compare}. The compared methods employ the same ResNet50 backbone. As we can see, the segmentation results of our method are more accurate than other methods, which is reflected in clearer object contours (first and fourth rows), more sensitive and accurate object detection (second and fifth rows), and less category mixing in segmentation map (third row). Compared with AugSeg~\cite{zhao2023augmentation} which is also based on two-branch architecture but with many color augmentations, our method can generate better results, demonstrating the effectiveness of the proposed augmentations for SSL Mars segmentation task.

\subsection{Limitations and Discussions}
\zjh{
Our method is efficiently designed in terms of data augmentation and pseudo-label optimization for semi-supervised Mars segmentation task. Notably, it scales well to the existing techniques, \eg, augmentation anchoring and distribution alignment in~\cite{berthelot2019remixmatch} and is simple to implement, making it a strong model to provide the basis for future work.

\revise{We present some failure cases in Fig.~\ref{fig:failure} As we can see, the limitations mainly lie in two aspects: 1) There is confusion between similar categories as shown in the first case. For example, to distinguish the rock and bedrock, the model needs to classify whether the rock is exposed to the ground, which can be difficult. One direction is to carry out specific designs, \eg, training a independent classifier, for these difficult categories. 2) The images directly taken by the rovers on Mars often have a long-tailed label distribution, which may affect the reliability of the pseudo-labels in semi-supervised learning, causing the noisy prediction or misclassification in the tail classes as shown in the second and third case. We point out these observed problems as future work, in the hope that more meaningful works will emerge.}
} 

\section{Conclusion}
\label{sec:conclusion}
In this paper, we address the SSL for Mars semantic segmentation problem from both data and method perspective. First, we propose a fine-grained annotated dataset S$^{5}$Mars for Martian terrain segmentation. This dataset provides sparse and high-confidence labeled data, which effectively assists the subsequent Mars exploration work. Then, we propose a simple yet effective SSL framework. Specifically, we analyze the effect of current used augmentations for Mars image segmentation. And two effective augmentations, \textit{AugIN} and \textit{SAM-Mix} are further proposed to improve the model performance. Meanwhile, a soft-to-hard consistency learning strategy is introduced to fully utilize the unlabeled data in a confidence-based manner. Extensive comparison and ablation experiments demonstrate the effectiveness of our method.

\small
\bibliography{sample-base}
\bibliographystyle{IEEEtran}

\begin{IEEEbiography}
[{\includegraphics[width=1in,height=1.25in,clip,keepaspectratio]{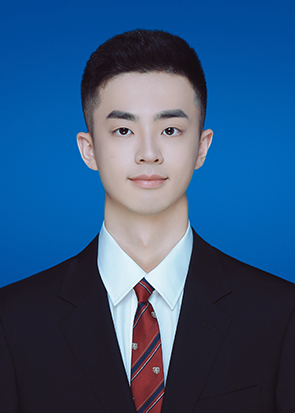}}]{Jiahang Zhang} received the B.S. degree in computer science from Peking University, Beijing, China, in 2023, where he is currently pursuing the Ph.D. degree with the Wangxuan Institute of Computer Technology. His current research interests include action recognition and self-supervised learning.
\end{IEEEbiography}

\vspace{5mm}

\begin{IEEEbiography}
[{\includegraphics[width=1in,height=1.25in,clip,keepaspectratio]{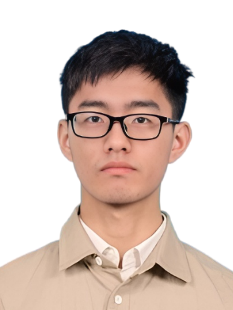}}]
{Lilang Lin}(Student Member, IEEE) received the B.S. degree in data science from Peking University, Beijing, China, in 2021, where he is currently pursuing the Ph.D. degree with the Wangxuan Institute of Computer Technology. His current research interests include action recognition, self-supervised learning, and unsupervised learning.
\end{IEEEbiography}

\vspace{5mm}

\begin{IEEEbiography}
[{\includegraphics[width=1in,height=1.25in,clip,keepaspectratio]{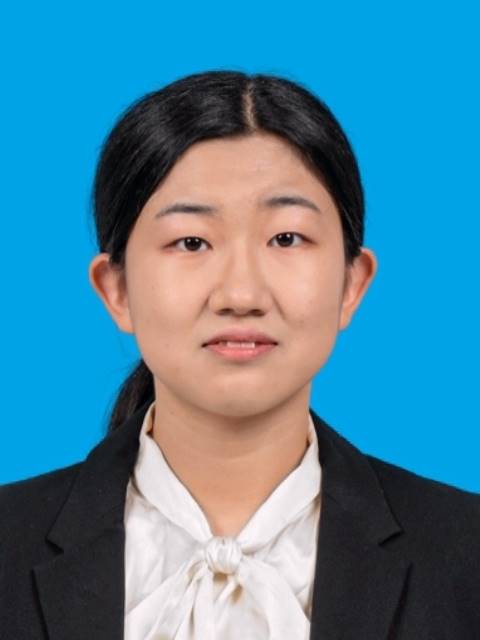}}]
{Zejia Fan} received the B.S. degree in Computer Science in 2021 from Peking University, Beijing, China, where she is currently working toward the Ph.D degree with the Wangxuan Institute of Computer Technology. Her current research interests include image enhancement and deep learning.
\end{IEEEbiography}

\vspace{5mm}

\begin{IEEEbiography}
[{\includegraphics[width=1in,height=1.25in,clip,keepaspectratio]{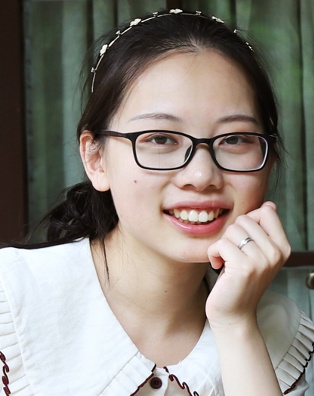}}]
{Wenjing Wang}(Student Member, IEEE) received the B.S. degree in data science from Peking University, Beijing, China, in 2019, where she is currently pursuing a doctoral degree with the Wangxuan Institute of Computer Technology, Peking University. 
She has authored over 20 technical articles in refereed journals and proceedings, and she holds five granted patents.
Her current research interests include image enhancement, image synthesis, and deep learning.
\end{IEEEbiography}

\vspace{5mm}

\begin{IEEEbiography}
[{\includegraphics[width=1in,height=1.25in,clip,keepaspectratio]{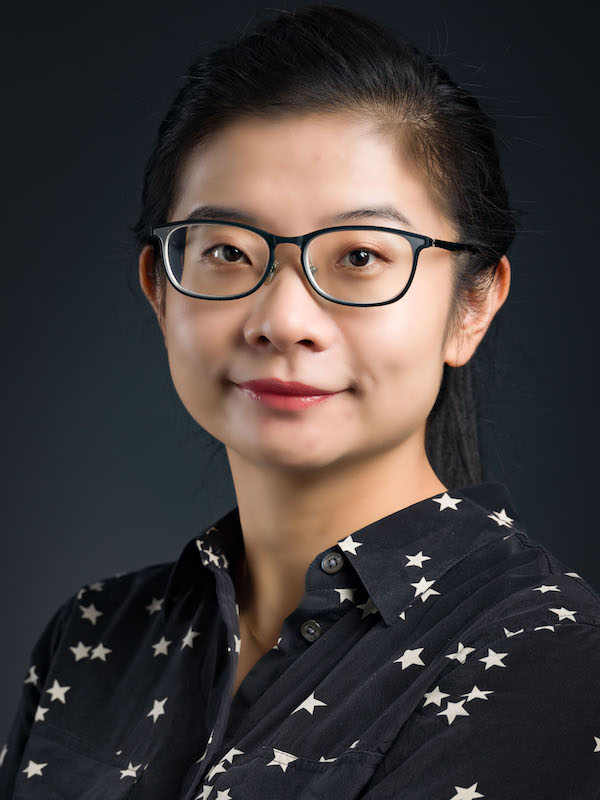}}]
{Jiaying Liu} (Senior Member, IEEE) received the PhD degree (Hons.) in computer science from Peking University, Beijing China, 2010. She is currently an Associate Professor, Boya Young Fellow with the Wangxuan Institute of Computer Technology, Peking University, China. She has authored more than 100 technical articles in refereed journals and proceedings, and holds 70 granted patents. Her current research interests include multimedia signal processing, compression, and computer vision. She is a senior member of IEEE/CSIG, and a distinguished member of CCF. She was a visiting scholar with the University of Southern California, Los Angeles, California, from 2007 to 2008. She was a visiting researcher with Microsoft Research Asia, in 2015 supported by the Star Track Young Faculties Award. 
Dr. Liu has served as a member of Multimedia Systems and Applications Technical Committee (MSA TC), and Visual Signal Processing and Communications Technical Committee (VSPC TC) in IEEE Circuits and Systems Society. She received the IEEE ICME 2020 Best Paper Award and IEEE MMSP 2015 Top10\% Paper Award. She has also served as the Associate Editor of the IEEE Trans. on Image Processing, the IEEE Trans. on Circuits Systems for Video Technology and Journal of Visual Communication and Image Representation, the Technical Program Chair of ACM MM Asia-2023/IEEE ICME-2021/ACM ICMR-2021/IEEE VCIP-2019, the Area Chair of CVPR-2021/ECCV-2020/ICCV-2019, ACM ICMR Steering Committee member and the CAS Representative at the ICME Steering Committee. She was the APSIPA Distinguished Lecturer (2016-2017).

\end{IEEEbiography}

\end{document}